\documentclass[11pt]{article}

\usepackage[final]{acl}

\usepackage{times}
\usepackage{latexsym}
\usepackage[T1]{fontenc}
\usepackage[utf8]{inputenc}
\usepackage{microtype}
\usepackage{inconsolata}
\usepackage{graphicx}
\usepackage{amsmath}
\usepackage{amssymb}
\usepackage{booktabs}
\usepackage{multirow}
\usepackage{algorithm}
\usepackage{algorithmic}
\usepackage{subcaption}
\usepackage{tikz}
\usepackage{verbatim}
\usepackage{mdframed}
\usepackage{orcidlink}
\usetikzlibrary{arrows.meta, positioning, shapes.geometric, fit, calc}

\title{APEX-EM: Non-Parametric Online Learning for Autonomous Agents \\ via Structured Procedural-Episodic Experience Replay}

\author{Pratyay Banerjee\,\orcidlink{0000-0001-5634-410X}, Masud Moshtaghi\,\orcidlink{0000-0002-1287-6697} and Ankit Chadha\,\orcidlink{0000-0002-5029-0136} \\
  Amazon, AGI / Sunnyvale, USA \\
  \texttt{pratyay, mmasud, ankitrc@amazon.com} \\}

\begin{document}
\maketitle

\begin{abstract}
LLM agents rerun full reasoning for every task, even one they solved moments earlier. We introduce \textbf{APEX-EM}, a non-parametric experience memory that stores complete procedural-episodic traces in a typed Procedural Knowledge Graph (PKG) and retrieves them through three channels: semantic search, structural-signature matching over abstract operation sequences, and graph traversal. A Plan-Retrieve-Generate-Iterate-Ingest (PRGII) workflow produces, quality-gates, and commits experiences, indexing both successes and failures so the agent learns what to reuse and what to avoid. No weights change during deployment.

We evaluate on five benchmarks: BigCodeBench, KGQAGen-10k, HLE, Lifelong Agent Bench, and ALFWorld. Because prior work uses different backbones, we base our claims on same-backbone comparisons that hold model capability fixed. On held-out BigCodeBench transfer with a shared GPT-4o backbone, APEX-EM gains +7.6\,pp over the no-memory baseline, $3.3\times$ MemRL's +2.3\,pp under the identical setup. On Lifelong Agent Bench with a shared GPT-4o-mini backbone, it gains +1.4\,pp (OS) and +1.0\,pp (DB) cumulative success. On KGQAGen-10k, frozen memory transfers to a blind 1{,}079-question test split at 73.7\% versus 42.0\% with no memory, approaching an oracle handed the ground-truth subgraph (84.9\%). Across three Opus scales the memory gain stays at +27 to +32\,pp, so it adds to model capability rather than substituting for it.

Component analysis shows no single mechanism dominates: teacher feedback is negligible for code but adds +10.3\,pp on structured queries, structural signatures give $3.3\times$ the transfer of semantic-only retrieval, and within-epoch iteration recovers most of the gain when rich feedback is unavailable. These results argue for modular memory composed per domain.

\end{abstract}

\section{Introduction}

Large language model (LLM)-based autonomous agents excel at multi-step reasoning, code generation, and tool use~\cite{yao2022react, wei2022chain}. Yet unlike reinforcement learning agents, which maintain experience replay buffers storing (state, action, reward, next\_state) tuples for training stability~\cite{lin1992experience}, LLM agents are fundamentally \emph{stateless at the procedural level}: each task invocation triggers the full reasoning pipeline from scratch, even when a near-identical task was successfully completed moments earlier.

MemRL~\cite{memrl2025} represents the most significant advance toward closing this gap, organizing memory into Intent-Experience-Utility triplets with learned Q-values and demonstrating that non-parametric experience accumulation can match or exceed methods requiring gradient updates. Three open challenges remain beyond MemRL's scope: experience representations that preserve procedural structure (planning steps, iteration history, error-outcome links) rather than LLM-generated summaries; explicit dual-outcome indexing that separates \emph{what succeeded} from \emph{what failed} rather than encoding both implicitly through a scalar Q-value; and retrieval mechanisms that transfer across domains sharing operational structure but no lexical overlap, precisely the regime where embedding similarity degrades (intra-HLE similarity of 0.186~\cite{memrl2025}).

We introduce \textbf{APEX-EM}, a framework that extends the non-parametric experience-driven paradigm. ``Non-parametric'' here means that no model weights are updated during deployment; the experience memory itself is the sole learned component, growing monotonically as the agent solves tasks, analogous to how a nearest-neighbor classifier improves as its reference set expands. Our contributions form a logical chain:

\begin{enumerate}
\item We formalize \emph{what} an experience should encode: a \textbf{Procedural Knowledge Graph (PKG)} storing complete procedural-episodic traces as replayable units, each tagged with a \emph{structural signature}, an abstract operation sequence (e.g., [entity\_resolution $\rightarrow$ temporal\_filter $\rightarrow$ aggregation]) enabling retrieval of procedurally analogous plans across surface-dissimilar domains.

\item We define \emph{how} experiences are generated and committed: the \textbf{PRGII workflow} (Plan-Retrieve-Generate-Iterate-Ingest), where a Teacher model provides multi-dimensional feedback (correctness, completeness, efficiency, error categorization) and a quality gate determines what enters the memory.

\item We show \emph{how} stored experiences are retrieved and applied: a \textbf{dual-outcome Experience Memory} with hybrid retrieval combining semantic search, structural signature matching, and graph traversal. Successful plans provide reusable procedure templates; failed plans provide structured error annotations (error class, root cause, recovery procedure), implementing the procedural equivalent of hindsight experience replay~\cite{andrychowicz2017hindsight}.
\end{enumerate}

We evaluate on five benchmarks (BigCodeBench~\cite{zhuo2025bigcodebench}, KGQAGen-10k~\cite{zhang2025kgqagen}, HLE~\cite{phan2025hle}, Lifelong Agent Bench~\cite{zheng2025lifelong}, and ALFWorld~\cite{shridhar2021alfworld}) using Claude Sonnet 4.5 and Opus 4.6 as primary backbones, with same-backbone comparisons (GPT-4o, GPT-4o-mini, GPT-5-mini) to control for model capability differences.

\section{Related Work}

\textbf{Memory for LLM Agents.} Reflexion~\cite{shinn2023reflexion} stores verbal reflections in episodic buffers; Voyager~\cite{wang2023voyager} builds verified code snippet libraries; MemGPT~\cite{packer2023memgpt} introduces OS-inspired memory hierarchies; Mem0~\cite{chhikara2025mem0} uses entity-centric relational graphs; MemP~\cite{fang2025memp} distills procedural knowledge. MemRL~\cite{memrl2025} advances beyond these by assigning learned Q-values to Intent-Experience-Utility triplets and re-ranking by composite similarity-utility scores. All store either unstructured reflections, code fragments, or flat summaries; none provides structured procedural traces with dual-outcome indexing and cross-domain structural retrieval. Retrieval-augmented generation (RAG) systems~\cite{lewis2020rag} retrieve documents for grounding but do not accumulate execution experience; classical plan libraries store reusable plans but lack LLM-based adaptation and structural signature matching. Closest in spirit, APEX-MEM~\cite{banerjee-etal-2026-apex} likewise uses a structured graph, organizing dialogue as a temporally grounded property graph for long-term \emph{conversational} recall, whereas APEX-EM accumulates procedural-episodic \emph{experience} for task execution.

\textbf{Experience Replay \& Online Learning.} APEX-EM is conceptually an experience replay buffer~\cite{lin1992experience} for LLM agents, but operates through in-context learning rather than gradient descent, stores high-level procedural plans rather than low-level transitions, and retrieves semantically and structurally rather than randomly. This positions it as a form of non-parametric experience-driven improvement~\cite{shalev2012online}: the agent improves during deployment as its reference set grows, analogous to a nearest-neighbor classifier, though without formal regret guarantees. MemRL is the closest prior work but stores LLM-summarized reflections rather than full execution traces, and its transfer degrades when semantic similarity is low.

\textbf{Case-Based Reasoning and Plan Abstraction.} The CBR cycle~\cite{aamodt1994cbr} (Retrieve, Reuse, Revise, Retain) bears structural resemblance to PRGII, and plan abstraction has a long history in classical AI planning: derivational analogy~\cite{veloso1994planning} reuses abstract proof structures across problems, and hierarchical task networks decompose tasks into reusable operator sequences~\cite{erol1994htn}. APEX-EM's structural signatures draw on this lineage. The key distinctions are: (1) extraction is performed by an LLM rather than hand-crafted rules, enabling open-vocabulary operation discovery without a predefined taxonomy; (2) dual-outcome indexing stores both successes and failures with structured error annotations (CBR typically retains only successes); and (3) retrieval combines structural matching with semantic and graph-based channels rather than relying on a single similarity metric. Full related work discussion in Appendix~\ref{app:related_work}.

Having established how APEX-EM relates to prior work, we now describe the framework in detail.

\section{Experience Replay Framework}

\subsection{Experience Representation}

Existing memory systems for LLM agents store flat text summaries~\cite{memrl2025}, code snippets~\cite{wang2023voyager}, or entity relationships~\cite{chhikara2025mem0}. These representations discard the decision-outcome chain linking planning choices to execution results. We propose a structured representation that preserves this chain, enabling retrieval by procedure, by error class, or by abstract operation pattern.

We consider an agent $\mathcal{A}$ receiving tasks $\{t_1, \ldots, t_n\}$ and producing executable artifacts $a_i$ evaluated by environment $\mathcal{E}$. The agent's policy is $a_i = \pi_\theta(t_i \mid \mathcal{M}^{(i-1)})$, where $\mathcal{M}^{(i-1)}$ denotes the Experience Memory accumulated from the first $i{-}1$ tasks; $\theta$ remains frozen throughout.

An \textbf{Experience} is the fundamental unit of knowledge: a structured record encoding the complete procedural-episodic trace of a task execution:
\begin{align}
\text{Exp} = \big(&\text{id}, \sigma, \mathcal{R}_G, \mathcal{R}_P, \mathcal{R}_E, \nonumber \\
&\mathcal{R}_T, \mathcal{R}_{err}, \mathcal{R}_{patch}, (c, \eta, \kappa, q, \mathcal{F}_T), s\big)
\end{align}
where $\sigma = [\text{op}_1, \ldots, \text{op}_K]$ is the \emph{structural signature}, an ordered sequence of abstract operations (e.g., entity\_resolution $\to$ schema\_traversal $\to$ temporal\_filter $\to$ aggregation), and $s \in \{\text{successful}, \text{failed}\}$ is determined by the quality gate ($s = \text{successful}$ iff $q \geq \theta$). The six reflection layers separate concerns: \textbf{Goal} ($\mathcal{R}_G$) captures what to achieve (task embedding, constraints, verification contract); \textbf{Procedure} ($\mathcal{R}_P$) captures how to execute (parameterized step-by-step template, portable across tasks); \textbf{Evidence} ($\mathcal{R}_E$) captures what was observed (content-addressed artifacts with provenance); \textbf{Execution Trace} ($\mathcal{R}_T$) archives iteration-by-iteration progress; \textbf{Error Registry} ($\mathcal{R}_{err}$) decomposes failures into error class, root cause, and recovery procedure; and \textbf{Patches} ($\mathcal{R}_{patch}$) encode reusable fixes as program deltas.

The \textbf{Execution Trace} ($\mathcal{R}_T$) preserves the full iteration history (intent, artifact, execution result, validation outcome, and feedback per iteration); failed attempts are archived with error analysis rather than discarded (per-field schema in Appendix~\ref{app:framework_details}).

The \textbf{Evaluation} layer scores each execution via a Teacher model on correctness $c$, efficiency $\eta$, and completeness $\kappa$, combined as $q = w_c \cdot c + w_\eta \cdot \eta + w_\kappa \cdot \kappa$ ($w_c = 0.9$, $w_\eta = w_\kappa = 0.05$). The permissive threshold $\theta = 0.3$ retains partially correct attempts (since $w_c=0.9$, it requires $c \geq 0.33$) so their procedural information is not lost. The Teacher's qualitative feedback $\mathcal{F}_T$ is decomposed into the Error Registry and Patch Reflections for failed executions, turning unstructured feedback into typed, queryable fields.

Each layer is independently queryable: an agent can retrieve by structural signature, by error class, by procedure template, or by entity reference. Both successful and failed experiences share the same schema; the distinction is the reward outcome from the quality gate, not a difference in record type. This is analogous to how RL replay buffers store transitions regardless of reward sign.

\subsection{Procedural Knowledge Graph}
\label{sec:pkg}

The Experience Memory $\mathcal{M}$ is organized as a typed graph, the Procedural Knowledge Graph (PKG), with five node types: \textbf{Entity} (real-world objects with properties), \textbf{Experience} (complete execution records), \textbf{Sub-Task} (decomposed sub-problems), \textbf{Operation} (abstract procedural steps forming structural signatures), and \textbf{TaskTopic} (category hubs). Ten edge types encode relationships: \texttt{FOLLOWED\_BY} (temporal ordering between Operations), \texttt{RELATES\_TO} (entity-property associations), \texttt{USES} (operation-property links), \texttt{EQUIVALENT\_TO} (cross-domain entity equivalence), \texttt{MEMBER\_OF} (topic membership), \texttt{uses\_entity} (experience-entity links enabling world-change propagation), \texttt{similar\_to} (cosine $> 0.85$), \texttt{structurally\_similar\_to} (LCS $\geq 0.6$ on signatures), \texttt{derived\_from} (guidance provenance), and \texttt{supersedes} (quality-based versioning).

The key innovation is \textbf{structural signature extraction}: when the agent solves a task, the underlying reasoning pattern is extracted as an ordered sequence of Operation nodes. The extraction process operates in three stages. First, the LLM-based DomainAdapter identifies abstract operations from the task description and execution trace (e.g., ``filter by date range'' $\to$ \texttt{temporal\_filter}). Second, each extracted operation is resolved against the PKG via the \textbf{EntityResolver}: the mention is embedded, vector-searched against existing Operation nodes of the same type, and mapped to an existing node if similarity exceeds threshold $\tau_r$ (with LLM disambiguation for ambiguous cases). Third, the resolved operations are ordered by execution dependency and linked via \texttt{FOLLOWED\_BY} edges, producing the structural signature $\sigma = [\text{op}_1, \ldots, \text{op}_K]$.

This resolution ensures structurally equivalent operations across tasks (e.g., ``group by category'' and ``aggregate by type'') map to the same Operation node, producing comparable signatures even when surface descriptions differ. The result enables cross-domain transfer: a plan for ``compare Curry's stats across seasons'' (entity\_resolution $\to$ temporal\_filter $\to$ aggregation $\to$ comparison) matches ``compare Tesla's revenue across quarters'' despite zero lexical overlap. Signature extraction is a side effect of graph construction; each task processed enriches the PKG, improving resolution quality for future tasks.

\subsection{PRGII Workflow}

The PRGII (Plan-Retrieve-Generate-Iterate-Ingest) workflow governs task execution through five phases. We introduce the workflow by tracing one task through all five, using the running example below.

\begin{mdframed}[linewidth=0.5pt,innertopmargin=4pt,innerbottommargin=4pt,innerleftmargin=6pt,innerrightmargin=6pt,backgroundcolor=black!2]
\small\textbf{Running example (KGQA).} Task $t$: \emph{``How many points did Stephen Curry score in the 2015--16 NBA season?''} Its hypothesized structural signature is $\hat{\sigma}=[\texttt{entity\_resolution}\to\texttt{schema\_traversal}\to\texttt{temporal\_filter}\to\texttt{aggregation}]$. We follow $t$ through each phase below.
\end{mdframed}

\begin{figure*}[t]
\centering
\begin{subfigure}[b]{\textwidth}
\centering
\resizebox{0.84\textwidth}{!}{%
\begin{tikzpicture}[
  font=\footnotesize,
  stage/.style={rectangle,rounded corners=3pt,draw=black!60,fill=blue!6,minimum height=1.0cm,minimum width=1.9cm,align=center,line width=0.6pt},
  sub/.style={rectangle,rounded corners=2pt,draw=black!45,fill=orange!8,minimum height=0.5cm,minimum width=2.0cm,align=center,font=\scriptsize},
  lm/.style={circle,draw=orange!80!black,fill=orange!35,minimum size=2.6mm,inner sep=0pt},
  arr/.style={-{Stealth[length=2.4mm]},line width=0.7pt,black!55},
]
\node[stage] (plan) at (0,0) {\textbf{Plan}};
\node[stage] (retr) at (3.4,0) {\textbf{Retrieve}\\[-2pt]{\scriptsize $P^{+}/P^{-}$}};
\node[stage] (gen)  at (6.8,0) {\textbf{Generate}};
\node[stage] (iter) at (10.2,0) {\textbf{Iterate}\\[-2pt]{\scriptsize validate $\mathcal{V}_d$}};
\node[stage] (ing)  at (13.6,0) {\textbf{Ingest}\\[-2pt]{\scriptsize gate $\to\mathcal{M}$}};
\node[sub] (tu) at (0,-1.35) {Task Understanding};
\node[sub] (ed) at (0,-2.05) {Entity Discovery};
\node[lm] at (tu.east) {};
\node[lm] at (ed.east) {};
\draw[arr,black!35] (tu.north) -- (plan.south);
\draw[arr,black!35] (ed.north) -- (tu.south);
\node[lm] at (plan.north east) {};
\node[lm] at (gen.north east) {};
\node[lm] at (iter.north east) {};
\node[lm] at (ing.north east) {};
\draw[arr] (plan)--(retr);
\draw[arr] (retr)--(gen);
\draw[arr] (gen)--(iter);
\draw[arr] (iter)--(ing);
\draw[arr] (iter.north) to[bend right=40] node[above,font=\scriptsize]{refine $\times J$} (gen.north);
\node[lm] (lg) at (12.0,1.5) {};
\node[right=1mm of lg,font=\scriptsize] {= LLM call};
\end{tikzpicture}}
\caption{The PRGII pipeline. Each model-calling stage carries an LLM marker (legend); Plan's two sub-steps are separate boxes; Generate and Iterate form a refinement loop (up to $J$).}
\label{fig:pipeline_a}
\end{subfigure}

\vspace{2pt}

\begin{subfigure}[b]{\textwidth}
\centering
\resizebox{0.60\textwidth}{!}{%
\begin{tikzpicture}[
  font=\footnotesize,
  nd/.style={rectangle,rounded corners=3pt,draw=black!55,minimum height=0.5cm,minimum width=1.7cm,align=center,font=\scriptsize},
  chan/.style={-{Stealth[length=2.4mm]},line width=0.8pt},
]
\node[nd,fill=blue!10] (ent) at (0,1.5) {Entity};
\node[nd,fill=green!10] (exp) at (0,0.75) {Experience};
\node[nd,fill=orange!14] (op) at (0,0) {Operation};
\node[nd,fill=red!10] (st) at (0,-0.75) {Sub-Task};
\node[nd,fill=purple!10] (tt) at (0,-1.5) {TaskTopic};
\node[draw=black!35,rounded corners=4pt,fit=(ent)(exp)(op)(st)(tt),inner sep=7pt,label=above:{\scriptsize\textbf{Procedural Knowledge Graph}}] (pkg) {};
\node[rectangle,rounded corners=3pt,draw=black!60,fill=blue!6,minimum height=1.0cm,minimum width=1.9cm] (retr) at (6.6,0) {\textbf{Retrieve}};
\draw[chan,blue!60!black] (pkg.east) to[out=25,in=150] node[above,font=\scriptsize,pos=0.55]{semantic} (retr.west);
\draw[chan,orange!85!black] (pkg.east) to[out=0,in=180] node[fill=white,inner sep=1pt,font=\scriptsize,pos=0.5]{structural signature} (retr.west);
\draw[chan,red!65] (pkg.east) to[out=-25,in=210] node[below,font=\scriptsize,pos=0.55]{graph traversal} (retr.west);
\end{tikzpicture}}
\caption{The typed PKG (five node types) feeds Retrieve through three channels: semantic search, structural-signature matching, and graph traversal.}
\label{fig:pipeline_b}
\end{subfigure}
\caption{The PRGII workflow. \textbf{(a)}~A task flows Plan\,$\to$\,Retrieve\,$\to$\,Generate\,$\leftrightarrow$\,Iterate\,$\to$\,Ingest; retrieved successes ($P^+$) and failures ($P^-$) condition Generate, and the quality gate commits to memory $\mathcal{M}$. \textbf{(b)}~The three retrieval channels over the Procedural Knowledge Graph.}
\label{fig:pipeline}
\end{figure*}

\textbf{Plan.} Given task $t$, produces $\mathcal{P}(t) = (\text{TU}, E, S, \hat{\sigma})$: task understanding (intent, constraints, output format), discovered entities resolved against the PKG via the EntityResolver, relevant schemas or API signatures from the environment, and the hypothesized structural signature. The signature $\hat{\sigma}$ is extracted \emph{before} retrieval, enabling structure-based search even when the task has no semantic overlap with stored experiences. Entity resolution during planning also enriches the PKG: new entities are added as nodes, and existing entities gain additional relationship edges. \emph{Running example.} The planner resolves ``Stephen Curry'' to the existing node Q352159, discovers the Wikidata properties \texttt{P54} (member of sports team) and \texttt{P1351} (points scored), and emits $\hat{\sigma}$ before any retrieval.

\textbf{Retrieve.} Searches $\mathcal{M}$ via three channels: (1) semantic search over task embeddings for surface-similar experiences; (2) structural signature matching via LCS ($\text{sim}_\sigma(\hat{\sigma}, \sigma_i) = |\text{LCS}(\hat{\sigma}, \sigma_i)| / \min(|\hat{\sigma}|, |\sigma_i|)$) for procedurally analogous experiences regardless of domain; and (3) PKG traversal via relational edges (\texttt{derived\_from}, \texttt{uses\_entity}, \texttt{structurally\_similar\_to}) up to two hops, surfacing related experiences invisible to vector search. Candidates are separated into successful ($P^+$) and failed ($P^-$) examples, then re-ranked by composite score (Eq.~\ref{eq:ranking}). \emph{Running example.} Structural matching surfaces a prior plan for ``LeBron's assists across seasons'' (identical $\hat{\sigma}$, different entity) as a template $P^+$, while a past run that queried the wrong property is retrieved as a guardrail $P^-$.

\textbf{Generate.} The LLM produces artifact $a$ conditioned on planning context and retrieved examples: $a = \text{LLM}(\text{TU}, E, S, P^+, P^-; \phi_d)$, where $\phi_d$ is the domain-specific prompt template. Successful plans provide reusable Procedure Reflections as adaptable templates; failed plans provide guardrails via their Error Registry (structured error decomposition) and Patch Reflections (program deltas preventing known failure modes). The dual conditioning implements the procedural equivalent of learning from both positive and negative demonstrations. \emph{Running example.} The model adapts the retrieved SPARQL template to Curry's QID and the 2015--16 season filter, and the $P^-$ guardrail steers it away from the previously mistaken property.

\textbf{Iterate} (agent self-verification)\textbf{.} The agent verifies its own work: a domain-specific Validator $\mathcal{V}_d$ evaluates each attempt, producing structured feedback $(v_j, f_j)$ where $v_j \in \{0, 1\}$ is the validation outcome and $f_j$ identifies root cause, fix target, and specific corrections. This is analogous to a student checking answers before submission. The loop runs up to $J$ iterations (3 for code, 10 for queries) with stuck-loop detection that identifies identical failures across consecutive iterations and injects divergent guidance. With memory, error analysis from failed plans makes self-verification more targeted: the agent knows what to look for because similar failures are documented in $P^-$. \emph{Running example.} The first query returns an empty result; the validator flags a wrong property path and the QueryPlanner swaps in the correct relation, after which execution returns the score.

\textbf{Ingest} (teacher evaluation and commit)\textbf{.} An independent Teacher model evaluates the completed trace, analogous to a teacher grading submitted work. It scores correctness ($w_c = 0.9$), efficiency, and completeness, producing $q = w_c \cdot c + w_\eta \cdot \eta + w_\kappa \cdot \kappa$. The quality gate routes: $s = \text{successful}$ if $q \geq \theta$ (default $\theta = 0.3$), else $\text{failed}$. For failed executions, the Teacher's qualitative feedback is decomposed into the Error Registry (error class, root cause, recovery procedure) and Patch Reflections (trigger signature, program delta, rationale). Both outcomes are committed to $\mathcal{M}$ with full reflection layers, closing the online learning loop. The first loop (Iterate) improves within-task execution; the second (Ingest) ensures memory quality and generates structured annotations that make failed experiences informative for future retrieval. The graph grows monotonically; no catastrophic forgetting, no retraining required. \emph{Running example.} The Teacher scores the corrected query as correct; with $q\geq\theta$ it commits as a successful Experience carrying its procedure template and $\hat{\sigma}$, while the empty-result first attempt is archived as a $P^-$ with an Error Registry entry (wrong property path) for future tasks.

\subsection{Hybrid Retrieval}

The three retrieval channels address complementary failure modes. \textbf{Semantic search} handles tasks similar to previously seen ones, the standard regime where embedding-based retrieval excels. \textbf{Structural signature matching} enables cross-domain transfer where embedding similarity fails, precisely the regime where MemRL's retrieval degrades. MemRL's own analysis confirms this: on HLE, intra-dataset semantic similarity is only 0.186~\cite{memrl2025}, meaning most task pairs share negligible embedding overlap. In this regime, MemRL's Two-Phase Retrieval (semantic similarity + Q-value re-ranking) cannot surface relevant experiences because the first phase filters them out. Structural signatures bypass this limitation entirely: two tasks with zero lexical overlap but identical reasoning patterns (e.g., both requiring entity\_resolution $\to$ temporal\_filter $\to$ aggregation) achieve $\text{sim}_\sigma = 1.0$. \textbf{PKG traversal} discovers related experiences through entity and operation connections invisible to vector search, enabling multi-hop reasoning about experience relationships.

Candidates from all three channels are merged, deduplicated, and re-ranked by composite score:
\begin{align}
\text{score}(p) = &\;\alpha \cdot \text{sim}_{\text{sem}} + \beta \cdot \text{sim}_\sigma + \gamma \cdot \text{prox}_G \nonumber \\
&+ \delta \cdot q_p + \epsilon \cdot \text{rec}(p) \label{eq:ranking}
\end{align}
where $\text{prox}_G$ is graph proximity (inverse hop distance), $q_p$ is quality score, and $\text{rec}(p)$ is recency. We set $(\alpha, \beta, \gamma, \delta, \epsilon) = (0.4, 0.3, 0.2, 0.1, 0.1)$, weighting semantic and structural similarity highest while using quality and recency as tie-breakers.

The \textbf{dual-outcome separation} matters for in-context learning: successful experiences ($P^+$, top-3) are formatted as adaptable procedure templates with parameterized variables (ENTITY, SOURCE, FILTER), while failed experiences ($P^-$, top-2) carry structured error annotations (failure mode such as \texttt{entity\_disambiguation}, root cause, recovery) and Patch Reflections encoding program deltas. The agent thus learns both what to do and what to avoid from the same store.

We now evaluate whether these design choices translate to empirical gains across five diverse benchmarks.

\section{Experiments}

We treat APEX-EM's components as a studied design space, and our central finding is that \emph{no single memory mechanism dominates}: teacher feedback is negligible for code but critical for structured queries (+10.3\,pp on KGQA), structural signatures drive cross-domain transfer, and iteration recovers most of the gain when rich feedback is unavailable. We target two deployment goals: transfer to unseen tasks, and faster improvement on recurring tasks under weak supervision. Because prior work uses different backbones, we hold model capability fixed with same-backbone comparisons wherever a matched baseline exists.

\subsection{Setup}

\begin{table*}[t]
\centering
\resizebox{\textwidth}{!}{
\begin{tabular}{@{}lcccccccccc@{}}
\toprule
& \multicolumn{2}{c}{\textbf{BCB}} & \multicolumn{2}{c}{\textbf{HLE}} & \multicolumn{2}{c}{\textbf{Lifelong-OS}} & \multicolumn{2}{c}{\textbf{Lifelong-DB}} & \multicolumn{2}{c}{\textbf{ALFWorld}} \\
\cmidrule(lr){2-3} \cmidrule(lr){4-5} \cmidrule(lr){6-7} \cmidrule(lr){8-9} \cmidrule(lr){10-11}
\textbf{Method} & \textbf{SR} & \textbf{CSR} & \textbf{SR} & \textbf{CSR} & \textbf{SR} & \textbf{CSR} & \textbf{SR} & \textbf{CSR} & \textbf{SR} & \textbf{CSR} \\
\midrule
\multicolumn{11}{l}{\textit{Cross-backbone reference: MemRL, each on its own backbone (training split, 10 epochs)}~\cite{memrl2025}} \\
\midrule
No Memory & 48.5$^a$ & -- & 35.7$^c$ & -- & 67.4$^b$ & -- & 86.0$^b$ & -- & 77.7$^d$ & -- \\
RAG & 47.9$^a$ & 53.0$^a$ & -- & -- & 69.0$^b$ & 70.0$^b$ & 91.4$^b$ & 91.6$^b$ & 90.7$^d$ & 93.5$^d$ \\
Mem0 & 53.0$^a$ & 57.7$^a$ & -- & -- & 67.0$^b$ & 70.2$^b$ & 92.0$^b$ & 92.6$^b$ & 89.4$^d$ & 96.9$^d$ \\
MemP & 57.8$^a$ & 60.2$^a$ & 52.8$^c$ & 58.2$^c$ & 73.6$^b$ & 74.2$^b$ & 96.0$^b$ & 96.6$^b$ & 88.5$^d$ & 91.9$^d$ \\
MemRL & 59.5$^a$ & 62.7$^a$ & 57.3$^c$ & 61.3$^c$ & 78.8$^b$ & 80.4$^b$ & 96.0$^b$ & 97.2$^b$ & 94.9$^d$ & 98.1$^d$ \\
\midrule
\multicolumn{11}{l}{\textit{APEX-EM (ours): Sonnet 4.5 / Opus 4.6 (training split, 10 epochs)}} \\
\midrule
No Memory & 53.9 ± 0.8 $^a$ & -- & 36.6 ± 0.7 $^f$ & -- & 72.0 ± 1.0$^e$ & -- & 94.4 ± 0.4 $^e$ & -- & 88.2 ± 0.5 $^e$ & -- \\
APEX-EM & \textbf{83.3 ± 0.9 }$^e$ & \textbf{84.0 ± 0.8}$^e$ & \textbf{61.0 ± 1.2}$^f$ & \textbf{64.0 ± 0.6}$^f$ & \textbf{78.8 ± 0.4}$^e$ & \textbf{82.0 ± 0.6}$^e$ & \textbf{97.8 ± 0.2}$^e$ & \textbf{98.8 ± 0.2}$^e$ & \textbf{99.7 ± 0.1}$^e$ & \textbf{99.96 ± 0.2}$^e$ \\
$\Delta$ vs.\ No Memory (same backbone) & +29.4 & -- & +24.4 & +27.4 & +6.8 & -- & +3.4 & -- & +11.5 & -- \\
\bottomrule
\end{tabular}}
\caption{Main results across five benchmarks (KGQA in Table~\ref{tab:kgqa_results}). \textbf{Split: training} (tasks recur across 10 epochs). SR = last-epoch success rate; CSR = cumulative success rate (solved at least once across 10 epochs). Rows are grouped into a cross-backbone MemRL reference block and the APEX-EM block; because each method runs on its own backbone, the vertical gap between blocks mixes backbone with framework and is \emph{not} a controlled comparison. For controlled, same-backbone comparisons against MemRL, see the held-out transfer results in Table~\ref{tab:transfer_results} (BCB +7.6 vs.\ MemRL +2.3 on a shared GPT-4o, i.e.\ $3.3\times$; Lifelong +1.4/+1.0 CSR on a shared GPT-4o-mini) and the per-benchmark tables in Appendix~\ref{app:per_benchmark}. MemRL's BCB SR is 59.5 here (training split); its blind held-out value is 50.8 (Table~\ref{tab:transfer_results}). Backbones: $^a$GPT-4o (MemRL's), $^b$GPT-4o-mini, $^c$Gemini-3-pro (discontinued), $^d$GPT-5-mini, $^e$Sonnet 4.5, $^f$Opus 4.6 ($\approx$ Gemini-3-pro baseline). Mean $\pm$ std over 5 seeds.}
\label{tab:main_results}
\end{table*}

\textbf{Models.} Claude Sonnet 4.5 (BCB, KGQA, Lifelong, ALFWorld), Claude Opus 4.6 (HLE primary; scaling across Opus 4.5--4.7 in Appendix). Entity extraction: Sonnet 4.6. Embeddings: Amazon Titan Embed v2 (1024-d). For direct comparison, we report MemRL's published results with GPT-4o (BCB), GPT-4o-mini (Lifelong), GPT-5-mini (ALFWorld), and Gemini-3-pro (HLE). Throughout, a \emph{same-backbone} comparison means APEX-EM and MemRL run on the identical backbone named in parentheses (GPT-4o for BCB, GPT-4o-mini for Lifelong, GPT-5-mini for ALFWorld); HLE has no same-backbone comparison because MemRL's Gemini-3-pro backbone was discontinued.

\textbf{Baselines.} Following~\citet{memrl2025}: (1) No Memory, (2) RAG, (3) Self-RAG~\cite{asai2023selfrag}, (4) Reflexion~\cite{shinn2023reflexion}, (5) Mem0~\cite{chhikara2025mem0}, (6) MemP~\cite{fang2025memp}, (7) MemRL~\cite{memrl2025}. All use frozen backbones under identical conditions.

\textbf{Metrics.} Last Epoch Success Rate (SR), Cumulative Success Rate (CSR), and LASM Accuracy (KGQA) following~\citet{memrl2025, zhang2025kgqagen}.

\textbf{Configuration.} Quality threshold $\theta = 0.3$; correctness weight $w_c = 0.9$; max iterations: 3 (BCB), 10 (KGQA, HLE); top-3 successful + top-2 failed retrieval; structural matching threshold $\tau_\sigma = 0.6$; retrieval weights per Eq.~\ref{eq:ranking}; 10 epochs; temperature 0. All results report the mean of 5 runs with different random seeds. Only A5 uses Opus 4.5 as Teacher; all other configurations and all same-backbone comparisons use the backbone itself as judge. Configs A1--R1 use flat semantic memory; S1/S2 add structural signature matching.

APEX-EM separates agent self-verification (Iterate) from teacher-evaluated commit (Ingest); the ablation in Table~\ref{tab:ablation} isolates their contributions (detailed analysis in Appendix~\ref{app:comparison_note}).

\subsection{Main Results}

\begin{table}[t]
\centering
\small
\begin{tabular}{@{}llc@{}}
\toprule
\textbf{Category} & \textbf{Method} & \textbf{LASM Acc.} \\ \midrule
\multicolumn{3}{l}{\textit{Pure LLMs~\cite{zhang2025kgqagen}}} \\
\midrule
& GPT-4o & 54.2 \\
& GPT-4.1 & 58.0 \\
\midrule
\multicolumn{3}{l}{\textit{KG-RAG Models~\cite{zhang2025kgqagen}}} \\
\midrule
& GCR (GPT-4o)$^\dagger$ & 54.0 \\
& ToG (GPT-4o) & 52.8 \\
& PoG (GPT-4o) & 53.5 \\
\midrule
\multicolumn{3}{l}{\textit{Oracle retrieval~\cite{zhang2025kgqagen}}} \\
\midrule
& GPT-4o w/ SP & 84.9 \\
\midrule
\multicolumn{3}{l}{\textit{APEX-EM (ours, Sonnet 4.5, 2K train)}} \\
\midrule
& No Memory (A0) & 41.3 ± 0.6 \\
& Semantic + rich + iter.\ (R1) & 85.7 ± 1.2\\
& Structural (S1) & 88.5 ± 0.9\\
& Sem.\ + Structural (S2) & 90.6  ± 0.8\\
& Sem.\ + Struct.\ + Opus (A5) & \textbf{91.2} ± 1.2\\
\midrule
\multicolumn{3}{l}{\textit{Transfer (1,079-q test split, frozen memory)}} \\
\midrule
& A0: No Memory & 42.0 ± 0.8\\
& R1: Semantic & 68.7 ± 0.4 \\
& S1: Structural & \textbf{73.7} ± 0.7 \\
& A5: Sem.\ + Struct.\ + Opus & 72.7 ± 1.2 \\
& R1: Semantic (GPT-4o) & 64.1 ± 0.8 \\
& S1: Structural (GPT-4o) & 67.1 ± 0.6 \\
& A5 (GPT-4o) & 66.2 ± 0.6 \\
\bottomrule
\end{tabular}
\caption{KGQAGen-10k results. Top: training split. Bottom: transfer to unseen test split (frozen memory). $\dagger$ = fine-tuned. ``w/ SP'' = oracle supporting subgraph.}
\label{tab:kgqa_results}
\end{table}

Table~\ref{tab:main_results} presents consolidated results across four benchmarks, with KGQA detailed separately in Table~\ref{tab:kgqa_results}.

\textbf{KGQA.} On the training split, APEX-EM reaches 91.2\% SR / 95.3\% CSR, a +49.9\,pp improvement over the 41.3\% baseline. Rather than read this recurring-task number against the single-shot oracle, which would not be a controlled comparison, we foreground the fair frozen-transfer result: on the blind 1{,}079-question test split, APEX-EM transfers at 73.7\% (S1) with no oracle access, approaching the oracle reference of 84.9\% (GPT-4o handed the ground-truth supporting subgraph; Table~\ref{tab:kgqa_results}). We therefore do not claim to surpass the oracle; the controlled takeaway is that \emph{accumulated procedural experience can substitute for oracle knowledge-graph access}, since APEX-EM must discover relevant entities through its own planning rather than receiving ground-truth subgraphs.

\textbf{BigCodeBench.} APEX-EM reaches 83.3\% SR / 84.0\% CSR (+29.4pp over the Sonnet 4.5 baseline). To control for backbone capability, we run APEX-EM on the same GPT-4o backbone used by MemRL: on transfer tasks (Table~\ref{tab:transfer_results}), APEX-EM achieves +7.6pp over the GPT-4o baseline, $3.3\times$ MemRL's +2.3pp gain, confirming that the improvement is framework-driven rather than backbone-driven. Even without the Opus judge, the Sonnet-only S2 configuration achieves 81.1\%, indicating that structural retrieval itself drives most of the gain.

\textbf{HLE.} We use HLE to test one claim only, that the memory gain is additive to backbone capability, not as a head-to-head against MemRL: its Gemini-3-pro backbone\footnote{Gemini-3-Pro was discontinued by Google in March 2026.} was discontinued, so no same-backbone comparison is possible. Across three Opus checkpoints the gain is stable and additive: Opus 4.5 +28.1\,pp, Opus 4.6 +27.4\,pp, and Opus 4.7 +31.8\,pp (reaching 71.2\% CSR; Appendix Table~\ref{tab:hle_scaling}), so memory adds to capability rather than compensating for a weak model. Two HLE settings are kept distinct: the runtime curves use MemRL's full 2{,}500-task set over 10 epochs (no transfer split), while ablations use a fixed 1{,}000-question random subset (seed 42) for tractability. The judge scores against the gold answer, but its feedback names \emph{where} the reasoning failed without revealing the solution: for a Spin-bordism task (gold $\mathbb{Z}^5$), the feedback flagged that the ``homology groups of $BG_2$ are computed incorrectly'' and directed recomputation, after which the agent corrected itself the next epoch (full traces in Appendix~\ref{app:judge_feedback}).

\textbf{Lifelong Agent Bench.} APEX-EM matches MemRL on OS (78.8\% Last SR) and exceeds it on DB (+1.8pp). MemRL achieves a larger relative gain on OS (+11.4pp from 67.4\%) compared to APEX-EM (+6.8pp from 72.0\%), reflecting diminishing returns from a stronger baseline. On the same GPT-4o-mini backbone, APEX-EM beats MemRL's SR by +1.4pp (OS) and +1.0pp (DB).

\textbf{ALFWorld.} APEX-EM achieves 99.7\% SR / 99.96\% CSR (+4.8pp over MemRL). On the same GPT-5-mini backbone, APEX-EM achieves 99.1\% SR in a single epoch (Appendix Table~\ref{tab:alfworld_results}), surpassing MemRL's 10-epoch result. The repetitive structure of household tasks makes this domain particularly amenable to procedural memory: once the agent learns the correct action sequence for a task type, it transfers reliably to all instances.

\begin{figure*}[t]
  \centering
  \includegraphics[width=0.32\textwidth]{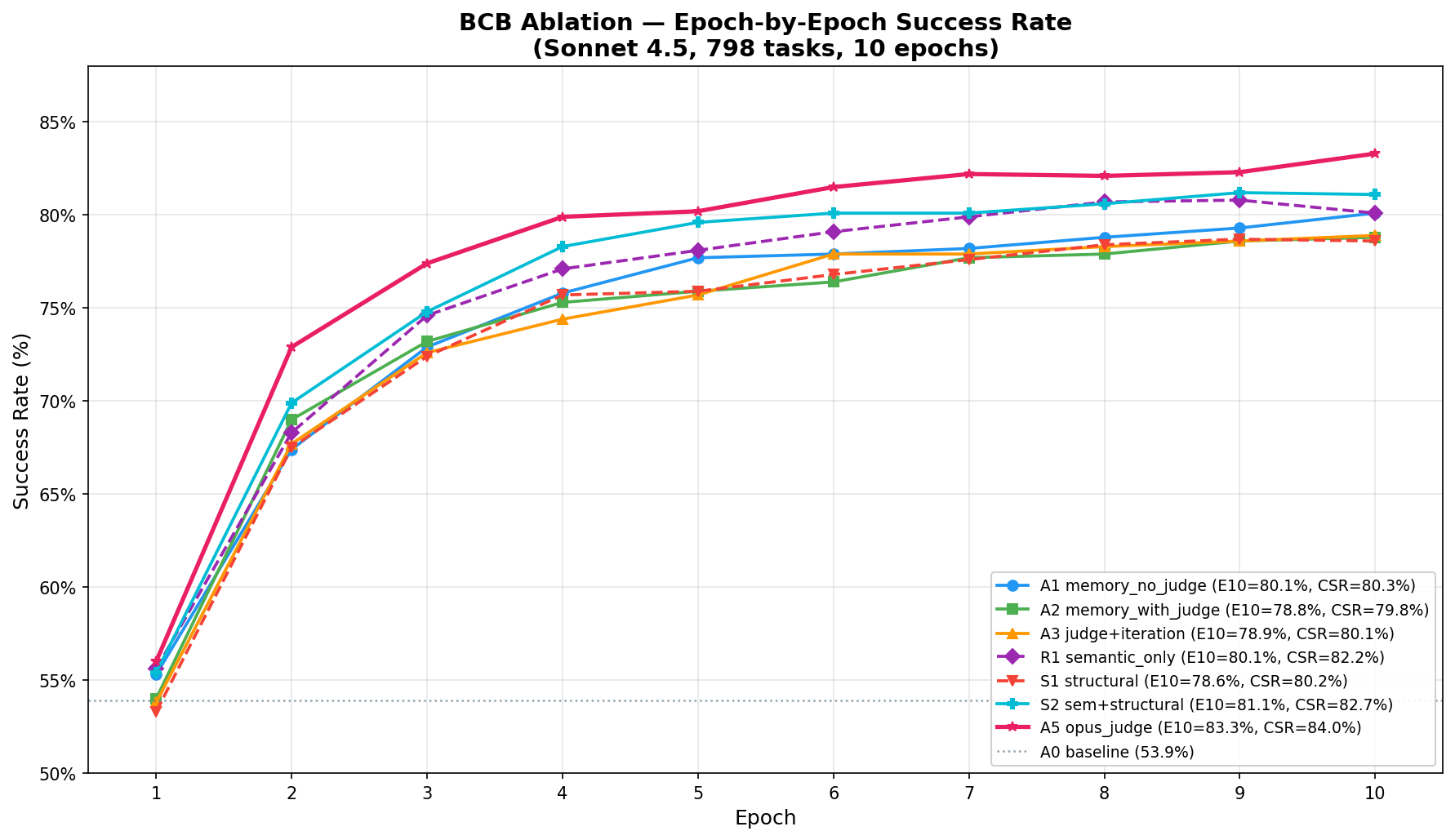}
  \hfill
  \includegraphics[width=0.32\textwidth]{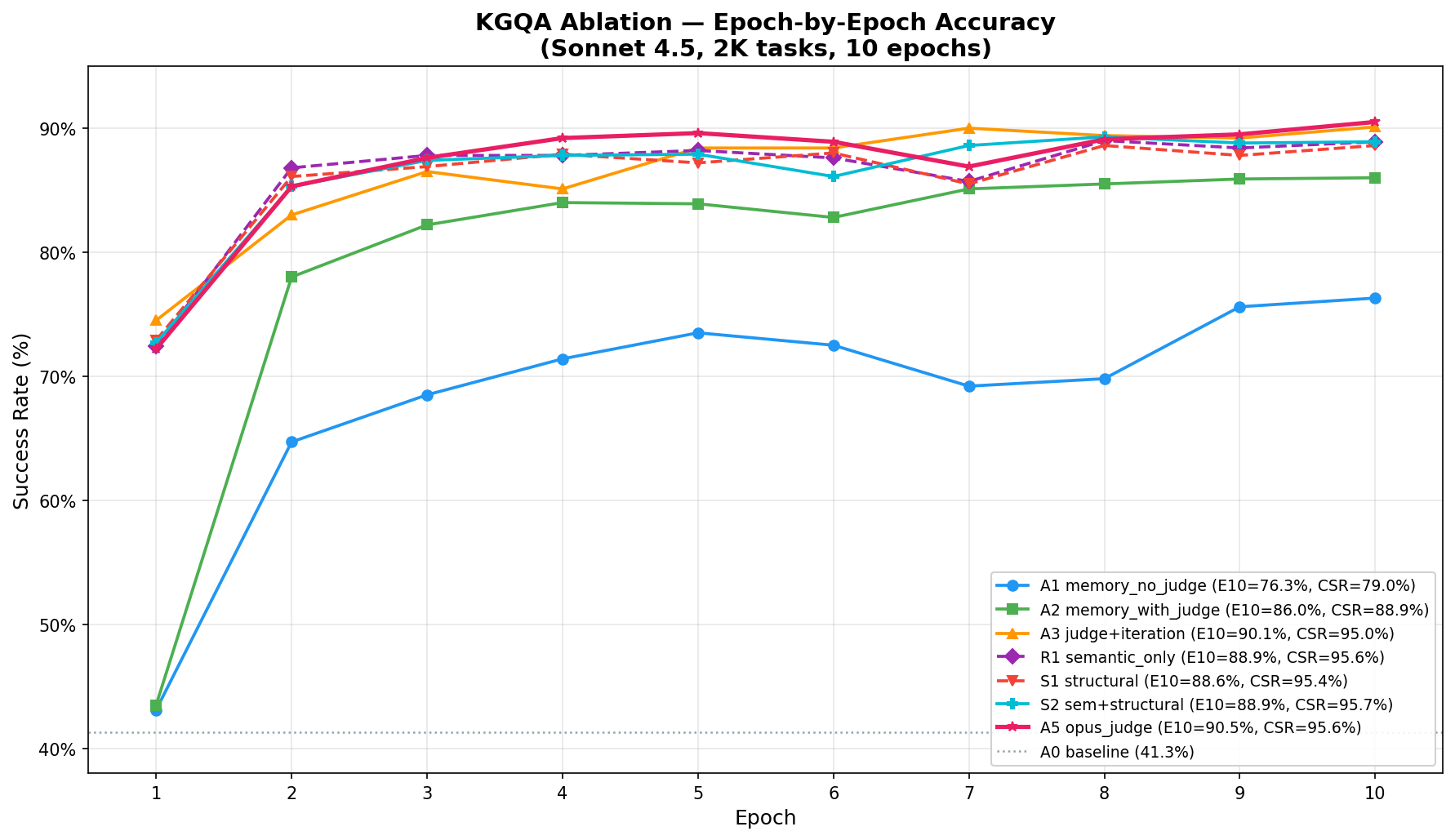}
  \hfill
  \includegraphics[width=0.32\textwidth]{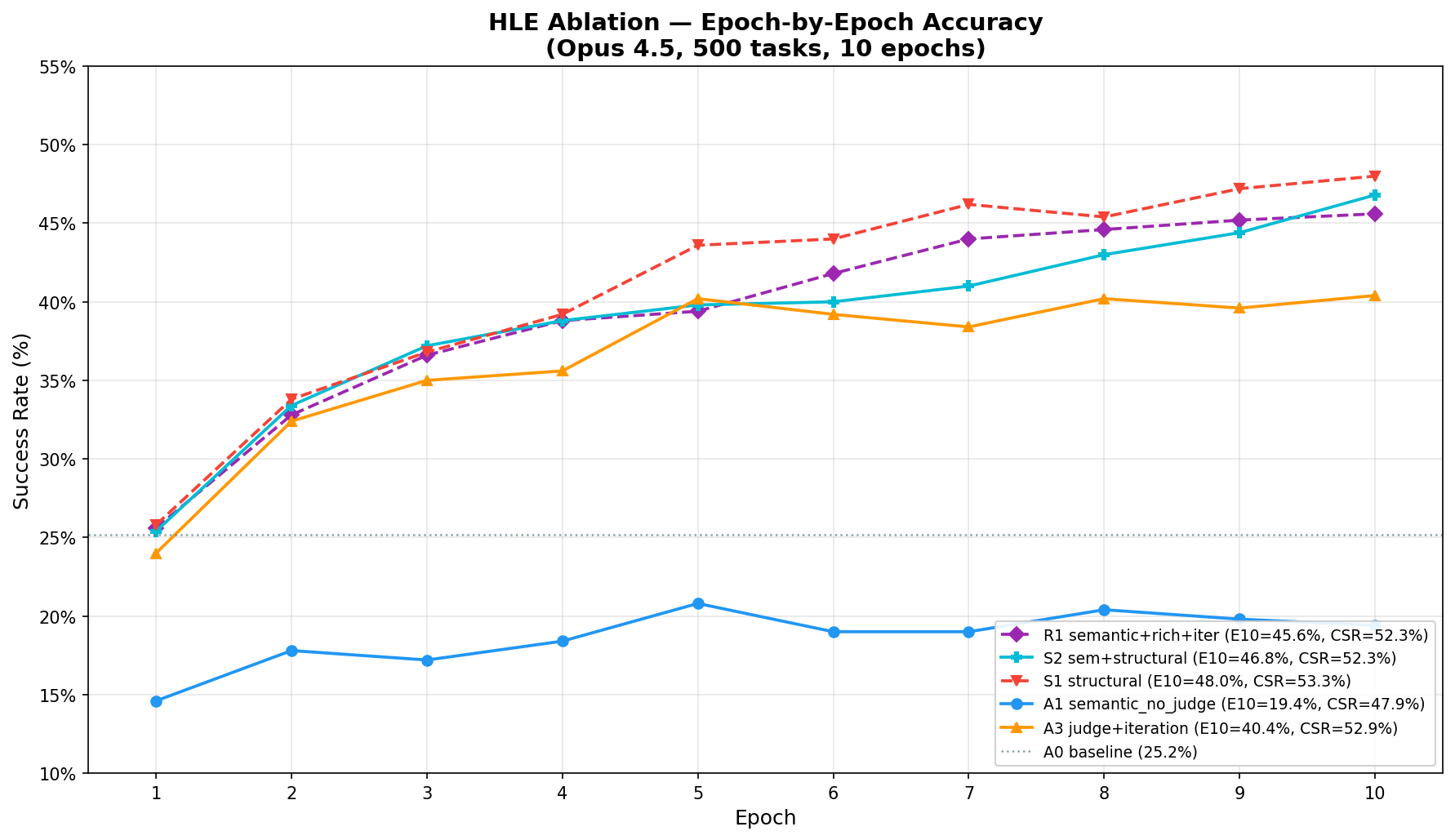}
  \caption{Epoch-by-epoch success rate. \textbf{Left:} BCB, A5 leads at 83.3\%. \textbf{Center:} KGQA, massive first-epoch jump from 41.3\% to $\sim$72--75\%; rich feedback (A2) reaches 85.9\% vs binary (A1) at 75.6\%. \textbf{Right:} HLE, S1 (structural) leads at 48.0\%; A1 (no judge) plateaus at $\sim$19\%.}
  \label{fig:learning_curve}
\end{figure*}

\subsection{Ablation Analysis}

\begin{table}[t]
\centering
\resizebox{\linewidth}{!}{
\begin{tabular}{@{}lcccc@{}}
\toprule
\textbf{Configuration} & \textbf{BCB SR} & \textbf{BCB CSR} & \textbf{KGQA SR} & \textbf{KGQA CSR} \\ \midrule
A0: No Memory (1 epoch) & 53.9\% ± 0.8 & -- & 41.3\% ± 1.2 & -- \\
A1: Semantic, no judge & 80.1\% ± 0.8 & 80.3\% ± 0.6 & 75.6\% ± 0.4 & 78.5\% ± 1.2 \\
A2: Semantic + judge & 78.8\% ± 0.4 & 79.8\% ± 0.9 & 85.9\% ± 1.0 & 88.8\% ± 0.4 \\
A3: Semantic + judge + iter. & 78.6\% ± 0.8 & 79.7\% ± 0.9 & 85.1\% ± 0.2 & 91.8\% ± 0.9 \\
R1: Semantic + rich + iter. & 80.1\% ± 0.2 & 82.2\% ± 0.4 & 85.7\% ± 0.8 & 94.8\% ± 0.6 \\
S1: Structural + rich + iter. & 78.6\% ± 0.6 & 80.2\% ± 0.8 & 88.5\% ± 0.7 & 94.8\% ± 0.4 \\
S2: Sem.\ + Struct.\ + rich + iter. & 81.1\%  ± 0.1 & 82.7\% ± 0.6 & 90.6\% ± 0.5 & 95.0\% ± 0.4 \\
A5: Sem.\ + Struct.\ + Opus judge & \textbf{83.3\%} ± 0.2 & \textbf{84.0\%} ± 0.4 & \textbf{91.2\%} ± 0.2 & \textbf{95.3\%} ± 0.4 \\
\bottomrule
\end{tabular}}
\caption{Ablation study on BigCodeBench and KGQAGen-10k. \textbf{Split: training} (tasks recur across 10 epochs). KGQA SR here measures last-epoch success rate on the training split (cf.\ LASM Accuracy in Table~\ref{tab:kgqa_results}, which uses the official evaluation metric on the blind test split).}
\label{tab:ablation}
\end{table}

Table~\ref{tab:ablation} isolates component contributions. Three key findings emerge:

\textbf{Judge feedback is task-dependent.} A1 slightly outperforms A2 on BCB (80.1\% vs 78.8\%), indicating that Teacher feedback provides no benefit for code generation, where execution results already provide unambiguous pass/fail signal. On KGQA, A2 exceeds A1 by +10.3pp (85.9\% vs 75.6\%); qualitative feedback explaining \emph{why} a SPARQL query failed is critical for structured query generation. The Teacher identifies specific relational errors (e.g., ``use P1346 instead of P166 for award winners; P166 is the `award received' property, but the question asks who \emph{won} the award, requiring the inverse relation P1346''), providing actionable corrections that binary pass/fail cannot convey. This finding has practical implications: expensive judge models are unnecessary for code generation but essential for structured queries.

\textbf{Iteration compensates for weak feedback.} A3 (iteration with binary signal) reaches 85.1\% on KGQA, nearly closing the gap to A2's 85.9\% rich feedback without iteration. The two mechanisms (cross-epoch memory retrieval and within-epoch self-correction) are complementary: memory provides the right starting point from prior experience, while iteration refines the current attempt through environmental feedback. When rich feedback is unavailable (e.g., proprietary APIs without detailed error messages), iteration provides an alternative path to high performance by allowing the agent to discover corrections through trial and error within a single task execution.

\textbf{Structural retrieval enables cross-domain gains.} S2 (combined retrieval) outperforms R1 (semantic-only) by +4.9pp SR on KGQA and +1.0pp on BCB, confirming that structural signatures provide an orthogonal retrieval axis. On KGQA, structural retrieval is particularly effective because query patterns recur across different entity types; a query about ``athletes who won awards in 2020'' shares the same structural signature as ``films that received nominations in 2019'' despite different entities and relations. S1 (structural-only) achieves 88.5\% on KGQA, exceeding semantic retrieval (R1: 85.7\%) by +2.8pp, while their combination (S2: 90.6\%) yields the largest gains. Appendix~\ref{app:channel_isolation} isolates each of the three channels individually (including graph traversal), and Appendix~\ref{app:sensitivity} sweeps the ranking weights and reports the minimal component set per domain.

\textbf{Iteration efficiency improves with memory.} Beyond accuracy, memory accumulation reduces computational cost: on KGQA (max 10 iterations), average iterations per task decrease by 37--43\% from epoch 1 to epoch 10, with A5 dropping from 4.40 to 2.50 iterations. On BCB (max 3 iterations), first-attempt solutions increase from 63.5\% to 76.8\%. This reveals the handoff from self-correction to memory-guided generation: in early epochs iteration does most of the work; in later epochs, retrieved experience makes the initial generation correct, rendering iteration unnecessary (cost and latency analysis in Appendix~\ref{app:latency}; per-domain failure modes in Appendix~\ref{app:failure_modes}).

\textbf{Memory gains scale with backbone capability.} Backbone scaling on HLE (Appendix Table~\ref{tab:hle_scaling}) reveals that stronger models benefit \emph{more} from procedural memory: Opus 4.5 gains +28.1pp, Opus 4.6 +27.4pp, and Opus 4.7 +31.8pp (reaching 71.2\% CSR). As model instruction-following ability improves, the agent becomes better at leveraging memory-guided in-context instructions: both the judge's qualitative feedback and retrieved procedure templates become more actionable. This suggests that APEX-EM's gains are not compensating for model weakness but are additive to capability, with a positive interaction between backbone strength and memory utility.

\subsection{Cross-Task Transfer}
\label{sec:transfer}

We freeze memory from 10 training epochs and evaluate on held-out data: 342 unseen BCB tasks, 1,079 unseen KGQA questions (Table~\ref{tab:kgqa_results}, bottom), and held-out Lifelong/ALFWorld splits. No new learning occurs during transfer.

\begin{table}[t]
\centering
\resizebox{\linewidth}{!}{
\begin{tabular}{@{}lccccc@{}}
\toprule
\textbf{Method} & \textbf{BCB} & \textbf{LLB-OS} & \textbf{LLB-DB} & \textbf{ALF} & \textbf{$\Delta$ BCB} \\
\midrule
\multicolumn{6}{l}{\textit{MemRL baselines$^\dagger$}} \\
\midrule
No Memory & 48.5\% & 67.4\% & 86.0\% & 77.7\% & --- \\
RAG & 47.9\% & 69.0\% & 91.4\% & 90.7\% & $-$0.6 \\
Mem0 & 48.5\% & 67.0\% & 92.0\% & 89.4\% & +0.0 \\
MemP & 49.4\% & 73.6\% & 96.0\% & 88.5\% & +0.9 \\
MemRL & 50.8\% & 74.6\% & 94.2\% & 97.9\% & +2.3 \\
\midrule
\multicolumn{6}{l}{\textit{APEX-EM (Ours, Sonnet 4.5)}} \\
\midrule
A0: No Memory & 45.9\% ± 0.2  & 74.7\% ± 1.2  & 93.3\% ± 0.8  & 84.3\% ± 1.0  & --- \\
A3: Sem.\ + iter. & 48.8\% ± 0.6  & 78.7\% ± 0.7  & 94.0\% ± 0.8  & 97.8\% ± 0.2  & +2.9 \\
R1: Sem.\ + rich & 50.3\% ± 0.5  & \textbf{79.3\%} ± 0.2  & 95.3\% ± 0.3  & 97.8\% ± 0.8  & +4.4 \\
S1: Structural & 51.2\% ± 0.2  & 78.0\% ± 0.8  & \textbf{96.7\%} ± 0.4  & 97.8\% ± 0.8  & +5.3 \\
A5: Full + Opus & \textbf{55.8\% ± 0.3 } & \textbf{79.3\% ± 0.4 } & \textbf{96.7\% ± 0.2 } & \textbf{100\% ± 0 } & \textbf{+9.9} \\
\midrule
\multicolumn{6}{l}{\textit{Same-backbone}} \\
\midrule
A5 (GPT-4o) & 53.5\% ± 0.4  & --- & --- & --- & +7.6 \\
A5 (GPT-4o-mini) & --- & 75.3\% ± 0.8  & 94.7\%  ± 0.4 & --- & --- \\
A5 (GPT-5-mini) & --- & --- & --- & 99.9\% ± 0.2  & --- \\
\bottomrule
\end{tabular}}
\caption{Cross-task transfer. \textbf{Split: held-out} (blind test tasks; memory frozen after 10 training epochs, no new learning). The \emph{Same-backbone} block is the controlled comparison to MemRL: APEX-EM (A5) and MemRL run on the identical backbone (GPT-4o for BCB, GPT-4o-mini for Lifelong, GPT-5-mini for ALFWorld). $\Delta$ BCB = pp over the no-memory baseline; on the shared GPT-4o backbone APEX-EM gains +7.6 vs.\ MemRL +2.3 ($3.3\times$). $^\dagger$MemRL~\cite{memrl2025} transfer results with respective backbones. KGQA transfer in Table~\ref{tab:kgqa_results} (S1: 73.7\% on test split).}
\label{tab:transfer_results}
\end{table}

Table~\ref{tab:transfer_results} reveals strong transfer with domain-dependent patterns. On BCB, A5 achieves +9.9pp over baseline; on the same GPT-4o backbone, +7.6pp, still $3.3\times$ larger than MemRL's +2.3pp. All MemRL baselines cluster near the no-memory baseline (47.9--50.8\%), confirming that semantic-only retrieval provides limited transfer for code generation. APEX-EM's structural signatures enable matching code patterns (e.g., data transformation pipelines) across tasks with different libraries and APIs.

On KGQA (Table~\ref{tab:kgqa_results}), \emph{iteration is required for transfer}: configs with iteration transfer at +29--32pp, while configs without show near-zero transfer (full ablation in Appendix). This demonstrates that retrieved experiences provide structural templates requiring adaptation to new entity structures. The iteration loop performs this adaptation by correcting entity bindings and relation paths through environmental feedback. S1 leads at 73.7\%, approaching the oracle bound (84.9\%) without oracle access.

On ALFWorld, A5 achieves \textbf{100\% transfer SR}: every held-out task solved on the first attempt with frozen memory. On Lifelong, retrieval strategy is task-dependent: R1 (semantic) leads on OS at 79.3\%, S1 (structural) leads on DB at 96.7\%, consistent with the finding that component value varies by domain.

\section{Conclusion}

We have presented APEX-EM, a framework enabling LLM agents to accumulate structured procedural experience without gradient updates. Same-backbone comparisons confirm gains are framework-driven ($3.3\times$ MemRL's transfer on BCB) and additive to backbone capability (+27--32pp across three scales). The central finding is that \emph{no single memory mechanism dominates}: teacher feedback, structural signatures, and self-verification each dominate in different task regimes. This argues for modular memory architectures composed per-domain.

Three directions follow: (1) \emph{learned retrieval policies} adapting weights per-task via Q-value utility estimation, unifying structural signatures with MemRL's utility scoring; (2) \emph{plan compression} distilling templates into compact programs as the PKG grows; and (3) \emph{automated signature discovery} from execution traces, removing domain-specific adapters.

\clearpage
\section*{Limitations}

The backbone models differ between APEX-EM (Claude Sonnet/Opus 4.5) and MemRL (GPT-4o/4o-mini/5-mini/Gemini-3-pro), complicating direct numerical comparison. We provide same-backbone comparisons on GPT-4o (BCB), GPT-4o-mini (Lifelong), and GPT-5-mini (ALFWorld) to control for model capability differences; only HLE lacks a same-backbone comparison due to the discontinuation of Gemini-3-pro, though the near-identical baselines (36.6\% vs 35.7\%) provide a controlled proxy.

The Teacher model (Opus 4.5) adds approximately 30\% computational overhead per task, and the full PRGII pipeline with iteration is substantially more expensive than single-shot inference. However, the Judge and memory ingestion phases are asynchronous and can run during \emph{dream time}, periods when GPUs have no active inference workload, decoupling memory creation cost from user-facing latency. In practice, this means the overhead is schedulable and does not impact serving throughput. The quality gate relies on either a Judge LLM (which may introduce evaluation noise) or human evaluators (which limits scalability).

The structural signature taxonomy is currently hand-designed; automated signature discovery from execution traces could improve coverage. The system does not provide formal convergence guarantees; the ``online learning'' characterization reflects the operational property of improvement through experience accumulation rather than theoretical regret bounds. All experiments rely on proprietary API-served models (Claude, GPT, Titan Embed) whose versions may be updated or deprecated over time, as evidenced by Gemini-3-pro's discontinuation during this work. While the 5-seed variance analysis and same-backbone controls mitigate version-specific artifacts, exact numerical reproduction may not be possible once model versions change.

\section*{Ethical Considerations}

The APEX-EM architecture stores complete execution traces, which may contain sensitive information from the task environment. Deployments must implement appropriate access controls and data retention policies. The dual-outcome indexing of failed plans requires careful handling to prevent storage of plans that failed due to privacy violations or harmful outputs. The offline-to-online pipeline, where plans generated by powerful models guide smaller models, raises questions about knowledge provenance and accountability.

Throughout the development of this work, generative AI systems were employed for language refinement and code test case generation. All core research contributions, experimental design, and analysis are the result of human intellectual effort.

\bibliography{custom}

\appendix
\section{Full Related Work}
\label{app:related_work}

\textbf{Memory Systems for LLM Agents.} The landscape of LLM agent memory has evolved rapidly. MemGPT~\cite{packer2023memgpt} pioneered OS-inspired memory hierarchies for managing context limitations. Reflexion~\cite{shinn2023reflexion} introduced verbal reinforcement learning through natural language reflections stored in episodic buffers. Voyager~\cite{wang2023voyager} built a skill library of verified code snippets for embodied agents. Generative Agents~\cite{park2023generative} demonstrated memory streams with periodic reflection and synthesis. More recently, A-MEM~\cite{xu2025amem} proposed Zettelkasten-inspired agentic memory with autonomous linking, Mem0~\cite{chhikara2025mem0} introduced entity-centric relational graphs, and MIRIX~\cite{wang2025mirix} achieved strong performance through six specialized memory stores with multi-agent routing. Zep~\cite{rasmussen2025zep} builds temporally-aware knowledge graphs for conversational memory. A comprehensive survey of memory mechanisms is provided by~\citet{zhang2024survey}. These systems store either unstructured reflections, code fragments, entity relationships, or narrative observations; none provides structured procedural traces with dual-outcome indexing and cross-domain structural retrieval.

\textbf{Experience Replay and Online Learning.} Experience replay~\cite{lin1992experience} is foundational in deep RL, storing past transitions for training stability. The APEX-EM is conceptually an experience replay memory/buffer for LLM agents, but with key differences: (a) experience replay updates weights via gradient descent; APEX-EM operates through in-context learning; (b) replay stores low-level (state, action, reward, next\_state) transitions; APEX-EM stores high-level procedural plans; (c) replay samples randomly or by priority; APEX-EM retrieves semantically and structurally; (d) replay does not distinguish positive from negative examples for different retrieval roles. This positions APEX-EM as a form of non-parametric online learning~\cite{shalev2012online}: the agent improves during deployment without weight updates, analogous to how a nearest-neighbor classifier improves as its reference set grows. The approach is model-agnostic, avoids catastrophic forgetting, and provides full auditability. MemRL~\cite{memrl2025} is the closest prior work: a non-parametric, frozen-backbone system that accumulates experience and improves at runtime. MemRL stores Intent-Experience-Utility triplets with learned Q-values and employs a Two-Phase Retrieval mechanism that first recalls candidates by semantic similarity, then re-ranks by Q-value. However, MemRL stores LLM-summarized reflections rather than full structured execution traces, uses Q-value-weighted ranking rather than structural signature matching, and implicitly encodes success/failure through Q-values rather than explicit dual-outcome indexing. Critically, MemRL's cross-domain transfer degrades when intra-dataset semantic similarity is low (as demonstrated in their ablation on HLE, where similarity of 0.186 reduced gains to memorization-like behavior), whereas APEX-EM's structural signatures are specifically designed for cross-domain transfer with zero lexical overlap.

\textbf{Case-Based Reasoning.} The CBR cycle~\cite{aamodt1994cbr, kolodner1993cbr} (Retrieve, Reuse, Revise, Retain) bears structural resemblance to PRGII. However, APEX-EM differs in maintaining dual-outcome indexing (CBR typically stores only successes), leveraging LLM-based in-context adaptation rather than hand-crafted transformation rules, using dense semantic embeddings and structural signatures rather than feature-based similarity, and employing quality-gated commitment with multi-criteria evaluation.

\textbf{Other Approaches.} ALMA~\cite{alma2025} uses meta-learning to discover memory designs but does not provide a concrete deployable architecture. SkillRL~\cite{skillrl2025} extracts reusable skills via RL weight updates rather than in-context learning. MemSkill~\cite{memskill2025} learns memory operations themselves rather than defining structured plan representations. LATS~\cite{zhou2023lats} combines LLM agents with Monte Carlo Tree Search but maintains experience only within single episodes without cross-task persistence.

\section{Datasets}
\label{app:datasets}

We evaluate APEX-EM on five benchmarks spanning code generation, structured query generation, multi-domain knowledge reasoning, OS interaction, and embodied navigation.

\textbf{BigCodeBench}~\cite{zhuo2025bigcodebench} comprises 1,140 tasks requiring composition of multiple library APIs to solve practical programming problems. Tasks demand understanding of library documentation, correct API composition across packages (e.g., \texttt{pandas}, \texttt{numpy}, \texttt{matplotlib}), and generation of complete, executable Python programs. Each task includes a comprehensive test suite; evaluation runs generated code through a sandboxed execution environment. We use the MemRL-compatible 7:3 train/transfer split (seed=42, 798 train / 342 transfer tasks).

\textbf{KGQAGen-10k}~\cite{zhang2025kgqagen} contains 10,787 natural language questions over Wikidata, constructed using LLM-guided subgraph expansion and SPARQL-based symbolic verification. Questions require multi-hop reasoning (2--5 hops), temporal filtering, comparative aggregation, and entity resolution across diverse topics. We evaluate on the standard test split of 1,079 examples and train on a 2,000-question sample.

\textbf{Humanity's Last Exam (HLE)}~\cite{phan2025hle} is a frontier benchmark of expert-level questions spanning mathematics, physics, chemistry, biology, computer science, engineering, humanities, and social sciences. Questions require deep domain expertise, multi-step reasoning, and creative problem-solving. We use a subset of HLE with Opus 4.5.

\textbf{Lifelong Agent Bench}~\cite{zheng2025lifelong} evaluates agents in a continuous learning setting involving Operating System (OS) and Database (DB) interactions, testing capacity to adapt to new tools and commands over a long horizon. We use 500 tasks/epoch for 10 epochs.

\textbf{ALFWorld}~\cite{shridhar2021alfworld} is an embodied navigation benchmark requiring agents to solve textual logic puzzles within a simulated household environment. We use 2,511 tasks/epoch for 3--5 epochs.

\section{Full Framework Details}
\label{app:framework_details}

\subsection{Experience Ontology}

\begin{table*}[t]
\centering
\caption{Condensed Experience ontology. Each Experience node in the PKG stores the following layered structure. Layers are independently queryable.}
\label{tab:experience-ontology}
\small
\resizebox{\textwidth}{!}{
\begin{tabular}{llll}
\toprule
\textbf{Layer} & \textbf{Field} & \textbf{Type} & \textbf{Description} \\
\midrule
\multirow{6}{*}{Goal Refl. $\mathcal{R}_G$}
& task\_description & String & Natural language task description \\
& task\_embedding & Vector[Float] & Embedding for semantic retrieval \\
& domain & Enum & web\_navigation $|$ code\_generation $|$ knowledge\_query $|$ \ldots \\
& constraints & Constraint[] & Budget limits, format requirements \\
& verification\_contract & Check[] & Predicates that must hold on completion \\
& goal\_signature & Object & Operations + structural\_signature + fingerprint \\
\midrule
\multirow{5}{*}{Proc.\ Refl.\ $\mathcal{R}_P$}
& procedure\_ref\_id & proc:\textlangle Name\textrangle:v\textlangle k\textrangle & Versioned template identifier \\
& params & Object & ENTITY, SOURCE, FILTER (template variables) \\
& steps[] & Step[] & Ordered ops with args and stop\_when conditions \\
& budgets & Object & max\_tool\_calls, max\_retries, max\_iterations \\
& checkpoints & Checkpoint[] & Verification gates between steps \\
\midrule
\multirow{4}{*}{Evid.\ Refl.\ $\mathcal{R}_E$}
& kind & Enum & web $|$ tool\_output $|$ db\_result $|$ file\_snapshot \\
& locator & Object & URL, tool name, query, timestamp \\
& content\_digest & Object & SHA-256 hash + byte count \\
& trust & Object & source\_type + authority\_score \\
\midrule
\multirow{5}{*}{Error Reg.\ $\mathcal{R}_{err}$}
& error\_class & Enum & constraint\_violation $|$ entity\_disambiguation $|$ tool\_failure $|$ \ldots \\
& occurred\_at & Object & step\_number + iteration\_number \\
& root\_cause & Object & hypothesis + confidence \\
& recovery\_procedure & Object & retry\_with\_patch $|$ fallback\_source $|$ escalate \\
& recovery\_outcome & Enum & recovered $|$ failed\_recovery $|$ escalated \\
\midrule
\multirow{4}{*}{Patch Refl.\ $\mathcal{R}_{patch}$}
& trigger\_signature & Object & symptoms + detected\_from patterns \\
& patch & Object & insert\_step $|$ replace\_logic + location + new\_logic \\
& rationale & String & Why this patch prevents the failure \\
& utility\_delta & Object & reliability and tool\_cost impact \\
\midrule
\multirow{4}{*}{Evaluation}
& correct, efficient, complete & Float & Scores 0--1 \\
& overall\_score $q$ & Float & Weighted composite \\
& teacher\_feedback $\mathcal{F}_T$ & String & Qualitative reasoning: why approach worked/failed \\
& status $s$ & Enum & successful $|$ failed \\
\bottomrule
\end{tabular}}
\end{table*}

\subsection{Entity Resolution}

When the Planning phase extracts potential entity mentions from a new task, each mention is resolved against the PKG via an \textbf{EntityResolver}:
\begin{enumerate}
\item Embed the mention text $m$ to obtain $\mathbf{e}_m$.
\item Vector search the PKG for candidates of the same node type and domain.
\item If the top candidate exceeds a similarity threshold $\tau_r$, map to the existing node (deduplication).
\item If multiple candidates are close, an LLM disambiguates.
\item Otherwise, create a new node and add it to the PKG.
\end{enumerate}
This ensures that structurally equivalent operations across tasks resolve to the same Operation node, producing comparable structural signatures even when surface descriptions differ.

\subsection{Structural Signature Extraction Quality}
\label{app:signature_quality}

We evaluate the reliability of LLM-based structural signature extraction using Sonnet 4.5 (the same model used for graph construction and entity resolution). On a held-out set of 200 tasks with human-annotated operation sequences across all five benchmarks, the DomainAdapter achieves $>$95\% F1 for operation extraction and entity resolution. To measure extraction consistency, we re-extract signatures from the same tasks across 5 independent runs (temperature 0): the mean pairwise LCS similarity between re-extracted signatures is 0.97, indicating near-deterministic extraction. Extraction failures (degenerate single-operation signatures) occur in $<$2\% of tasks and are handled gracefully: such tasks fall back to semantic-only retrieval.

\subsection{Domain-Specific Pipeline Instantiations}
\label{app:pipelines}

\paragraph{BigCodeBench (Code Generation).}
The BCB pipeline uses: \textbf{Plan}: \texttt{CodeDomainAdapter} extracts library names, function signatures, import requirements, and procedural steps resolved against the Entity Graph. \textbf{Generate}: produces structured JSON with \texttt{code} and \texttt{tests} fields; co-generated tests enable student self-testing. \textbf{Iterate}: \texttt{CodeExecutionValidator} runs code + tests in a 30-second sandboxed subprocess with structured feedback (error classification, root cause, fix target). \textbf{Ingest}: wraps the official \texttt{untrusted\_check} harness; correctness is binary.

\paragraph{KGQAGen-10k (Structured Query Generation).}
KGQA uses: \textbf{Plan}: \texttt{WikidataEntityLinker} performs entity linking with LLM-guided disambiguation and property discovery via Wikidata API. \textbf{Generate}: produces SPARQL queries with entity QIDs and property PIDs. \textbf{Iterate}: SPARQL execution against Wikidata; a \texttt{QueryPlanner} LLM analyzes results and proposes refined searches. \textbf{Ingest}: LLM-as-Judge comparing generated SPARQL against ground-truth.

\paragraph{HLE (Multi-Domain Knowledge Reasoning).}
HLE uses: \textbf{Plan}: \texttt{HLEDomainAdapter} extracts domain concepts and reasoning patterns. \textbf{Generate}: supports multimodal generation for image-based questions; retrieved experiences formatted as ``CRITICAL --- PREVIOUS ATTEMPTS''. \textbf{Iterate}: \texttt{HLEIterationController} with 7-action space: decompose, search, compute, verify, fact\_check, reformulate, finalize. \textbf{Ingest}: exact-match or LLM comparison (no answer leakage).

\subsection{Memory Maintenance}

As $\mathcal{M}$ grows, maintenance mechanisms ensure retrieval quality: (1) \textbf{Entity Swaps}: world knowledge changes propagate automatically via typed edges; (2) \textbf{Continuous Updates}: every quality-gated trace commits without duplicating structure; (3) \textbf{Compaction}: archives dominated experiences via \texttt{supersedes} edges, consolidates similar experiences into parameterized templates, and flags stale entities for re-validation.

\section{Per-Benchmark Results}
\label{app:per_benchmark}

\subsection{Code Generation (BigCodeBench)}

\begin{table}[t]
\centering
\resizebox{\linewidth}{!}{
\begin{tabular}{@{}llcc@{}}
\toprule
\textbf{Backbone} & \textbf{Method} & \textbf{Last SR} & \textbf{CSR} \\ \midrule
\multicolumn{4}{l}{\textit{MemRL baselines (GPT-4o)~\cite{memrl2025}}} \\
\midrule
GPT-4o & No Memory & 0.485 & -- \\
GPT-4o & Pass@10 & -- & 0.577 \\
GPT-4o & Reflexion & -- & 0.582 \\
GPT-4o & RAG & 0.479 & 0.530 \\
GPT-4o & Self-RAG & 0.500 & 0.558 \\
GPT-4o & Mem0 & 0.530 & 0.577 \\
GPT-4o & MemP & 0.578 & 0.602 \\
GPT-4o & MemRL & 0.595 & 0.627 \\
\midrule
\multicolumn{4}{l}{\textit{APEX-EM (ours, Claude Sonnet 4.5, GPT-4o)}} \\
\midrule
Sonnet 4.5 & No Memory (A0) & 0.539 & -- \\
Sonnet 4.5 & Semantic + Structural (S2, E10) & 0.811 & 0.827 \\
Sonnet 4.5 & Sem.\ + Struct.\ + Opus (A5, E10) & \textbf{0.833} & \textbf{0.840} \\
GPT-4o & Sem.\ + Struct.\ + Opus (A5, E10) & 0.811 & 0.815 \\
\bottomrule
\end{tabular}}
\caption{Code generation results on BigCodeBench (Full, 7:3 train/transfer split, seed=42). Last Epoch Success Rate / Cumulative Success Rate on train split. MemRL baselines from~\citet{memrl2025} with GPT-4o backbone. All APEX-EM configs complete (10 epochs).}
\label{tab:bcb_results}
\end{table}

Without memory, Sonnet 4.5 achieves 53.9\% (A0), comparable to GPT-4o's 48.5\%. With the full PRGII pipeline and Opus judge (A5), APEX-EM reaches 83.3\% SR / 84.0\% CSR, a +29.4pp improvement over the no-memory baseline, substantially exceeding MemRL's +11.0pp gain (48.5\%$\to$59.5\% SR, 62.7\% CSR). Without the Opus judge, S2 achieves 81.1\% / 82.7\% CSR.

\subsection{Structured Query Generation (KGQAGen-10k)}

\begin{table}[t]
\centering
\resizebox{\linewidth}{!}{
\begin{tabular}{@{}llcc@{}}
\toprule
\textbf{Category} & \textbf{Method} & \textbf{LASM Acc.} & \textbf{CSR} \\ \midrule
\multicolumn{4}{l}{\textit{Pure LLMs (no retrieval)~\cite{zhang2025kgqagen}}} \\
\midrule
 & LLaMA2-7B & 11.6 & -- \\
 & LLaMA-3.1-8B-Instruct & 30.5 & -- \\
 & Mistral-7B-Instruct-v0.2 & 22.8 & -- \\
 & GPT-4o-mini & 44.9 & -- \\
 & GPT-4 & 48.7 & -- \\
 & DeepSeek-Chat & 50.1 & -- \\
 & GPT-4o & 54.2 & -- \\
 & GPT-4.1 & 58.0 & -- \\
\midrule
\multicolumn{4}{l}{\textit{KG-RAG Models~\cite{zhang2025kgqagen}}} \\
\midrule
 & RoG (LLaMA2-7B)$^\dagger$ & 21.3 & -- \\
 & GCR (LLaMA-3.1 + GPT-4o)$^\dagger$ & 54.0 & -- \\
 & ToG (GPT-4o) & 52.8 & -- \\
 & PoG (GPT-4o) & 53.5 & -- \\
\midrule
\multicolumn{4}{l}{\textit{LLM with Supporting Subgraph (oracle retrieval)~\cite{zhang2025kgqagen}}} \\
\midrule
 & LLaMA2-7B (w/ SP) & 73.8 & -- \\
 & GPT-4o (w/ SP) & 84.9 & -- \\
\midrule
\multicolumn{4}{l}{\textit{APEX-EM (ours, Claude Sonnet 4.5)}} \\
\midrule
 & No Memory (A0, 1 epoch) & 41.3 & -- \\
 & Semantic + rich + iter.\ (R1) & 85.7 & 94.8 \\
 & Structural (S1) & 88.5 & 94.8 \\
 & Sem.\ + Structural (S2) & 90.6 & 95.0 \\
 & Sem.\ + Struct.\ + Opus (A5) & \textbf{91.2} & \textbf{95.3} \\
\midrule
\multicolumn{4}{l}{\textit{APEX-EM (ours, GPT-4o)}} \\
\midrule
  & Sem.\ + Struct.\ + Opus (A5) & 89.8 & 90.2 \\
\bottomrule
\end{tabular}}
\caption{Results on KGQAGen-10k~\cite{zhang2025kgqagen}. LASM Accuracy (\%) and Cumulative Success Rate (\%) on the training split. Baselines from~\citet{zhang2025kgqagen}; $\dagger$ = fine-tuned.}
\label{tab:kgqagen_results}
\end{table}

Without memory, Sonnet 4.5 achieves 41.3\% (A0, single epoch, no iteration). With the full PRGII pipeline and Opus judge (A5), APEX-EM reaches 91.2\% SR / 95.3\% CSR on the training split, a +49.9pp improvement over the no-memory baseline. As discussed in \S\ref{sec:transfer}, we do not read this recurring-task number against the single-shot oracle; the fair comparison is the blind-transfer result (S1 73.7\% vs oracle 84.9\%), and the takeaway is that accumulated experience can substitute for oracle KG access, since APEX-EM discovers entities through its own planning with no oracle access.

\subsection{Multi-Domain Knowledge Reasoning (HLE)}

\begin{table}[t]
\centering
\resizebox{\linewidth}{!}{
\begin{tabular}{@{}llcc@{}}
\toprule
\textbf{Backbone} & \textbf{Method} & \textbf{Last SR} & \textbf{CSR} \\ \midrule
\multicolumn{4}{l}{\textit{MemRL baselines (Gemini-3-pro$^*$)~\cite{memrl2025}}} \\
\midrule
Gemini-3-pro & No Memory & 0.357 & -- \\
Gemini-3-pro & Pass@10 & -- & 0.524 \\
Gemini-3-pro & Reflexion & -- & 0.530 \\
Gemini-3-pro & RAG & 0.500 & 0.548 \\
Gemini-3-pro & Self-RAG & 0.488 & 0.548 \\
Gemini-3-pro & Mem0 & 0.512 & 0.560 \\
Gemini-3-pro & MemP & 0.528 & 0.582 \\
Gemini-3-pro & MemRL & 0.573 & 0.613 \\
\midrule
\multicolumn{4}{l}{\textit{APEX-EM (ours, Claude Opus 4.5)}} \\
\midrule
Opus 4.5 & No Memory (A0) & 0.252 & -- \\
Opus 4.5 & Semantic, no judge (A1) & 0.194 & 0.479 \\
Opus 4.5 & Semantic + judge + iter.\ (A3) & 0.404 & 0.529 \\
Opus 4.5 & Semantic + rich + iter.\ (R1) & 0.456 & 0.523 \\
Opus 4.5 & Sem.\ + Structural (S2) & 0.468 & 0.523 \\
Opus 4.5 & Structural (S1) & 0.480 & 0.533 \\
\midrule
\multicolumn{4}{l}{\textit{APEX-EM backbone scaling (comparable to Gemini-3-pro)}} \\
\midrule
Opus 4.6 & No Memory (A0) & 0.366 & -- \\
Opus 4.6 & Structural (S1) & 0.610 & 0.640 \\
Opus 4.7 & No Memory (A0) & 0.394 & -- \\
Opus 4.7 & Structural (S1) & 0.684 & \textbf{0.712} \\
\bottomrule
\end{tabular}}
\caption{Multi-domain knowledge reasoning results on HLE~\cite{phan2025hle} (10 epochs). MemRL baselines from~\citet{memrl2025}. $^*$Gemini-3-pro discontinued March 2026. Opus 4.6/4.7 baselines (36.6\%/39.4\%) are comparable to Gemini-3-pro (35.7\%). APEX-EM with Opus 4.7 achieves 71.2\% CSR, exceeding MemRL's 61.3\% by +9.9pp.}
\label{tab:hle_results}
\end{table}

Without memory, Opus 4.5 achieves 25.2\% (A0). Structural retrieval (S1) leads at 48.0\% / 53.3\% CSR (+22.8pp over A0), with combined retrieval (S2: 46.8\% / 52.3\%) and semantic-only (R1: 45.6\% / 52.3\%) close behind. Memory without judge or iteration (A1) shows limited learning: 14.6\%$\to$19.4\% SR over 10 epochs, but achieves high CSR (47.9\%), indicating broad exploration without convergence. The judge feedback does not leak ground-truth answers; it identifies where reasoning went wrong (e.g., ``homology groups computed incorrectly'') without revealing the target answer.

\subsection{OS and Database Interaction (Lifelong Agent Bench)}

\begin{table}[t]
\centering
\resizebox{\linewidth}{!}{
\begin{tabular}{@{}llcccc@{}}
\toprule
& & \multicolumn{2}{c}{\textbf{OS Task}} & \multicolumn{2}{c}{\textbf{DB Task}} \\
\cmidrule(lr){3-4} \cmidrule(lr){5-6}
\textbf{Backbone} & \textbf{Method} & \textbf{Last SR} & \textbf{CSR} & \textbf{Last SR} & \textbf{CSR} \\ \midrule
\multicolumn{6}{l}{\textit{MemRL baselines (GPT-4o-mini)~\cite{memrl2025}}} \\
\midrule
GPT-4o-mini & No Memory & 0.674 & -- & 0.860 & -- \\
GPT-4o-mini & Pass@10 & -- & 0.756 & -- & 0.928 \\
GPT-4o-mini & RAG & 0.690 & 0.700 & 0.914 & 0.916 \\
GPT-4o-mini & Self-RAG & 0.646 & 0.732 & 0.891 & 0.898 \\
GPT-4o-mini & Mem0 & 0.670 & 0.702 & 0.920 & 0.926 \\
GPT-4o-mini & MemP & 0.736 & 0.742 & 0.960 & 0.966 \\
GPT-4o-mini & MemRL & 0.788 & 0.804 & 0.960 & 0.972 \\
\midrule
\multicolumn{6}{l}{\textit{APEX-EM (ours, Claude Sonnet 4.5)}} \\
\midrule
Sonnet 4.5 & No Memory (A0) & 0.720 & 0.740 & 0.944 & 0.950 \\
Sonnet 4.5 & Semantic + judge (A2) & 0.750 & 0.812 & 0.956 & 0.956 \\
Sonnet 4.5 & Semantic + judge + iter.\ (A3) & 0.786 & 0.812 & 0.972 & 0.982 \\
Sonnet 4.5 & Structural (S1) & 0.770 & 0.814 & 0.972 & 0.988 \\
Sonnet 4.5 & Sem.\ + Struct.\ + Opus (A5) & 0.780 & 0.806 & 0.956 & 0.986 \\
Sonnet 4.5 & Semantic + Structural (S2) & 0.788 & \textbf{0.820} & \textbf{0.978} & \textbf{0.988} \\
\midrule
\multicolumn{6}{l}{\textit{APEX-EM (ours, GPT-4o-mini) — same backbone as MemRL}} \\
\midrule
GPT-4o-mini & APEX-EM A5 & 0.788 & 0.818 & 0.963 & 0.982 \\
\bottomrule
\end{tabular}}
\caption{Lifelong Agent Bench~\cite{zheng2025lifelong} results: OS interaction and database tasks (500 tasks/epoch, 10 epochs). MemRL baselines from~\citet{memrl2025} with GPT-4o-mini backbone. Last SR = final epoch success rate; CSR = cumulative success rate (unique tasks solved at least once across all epochs). APEX-EM with GPT-4o-mini backbone beats MemRL on CSR (+1.4pp OS, +1.0pp DB) with same backbone.}
\label{tab:lifelong_results}
\end{table}

MemRL achieves 78.8\% (OS) and 96.0\% (DB) Last SR after 10 epochs with GPT-4o-mini. APEX-EM with Sonnet 4.5 achieves 78.8\% / 82.0\% CSR (OS) and 97.8\% / 98.8\% CSR (DB) with S2, matching MemRL's Last SR on OS while substantially exceeding it on DB (+1.8pp). On the same GPT-4o-mini backbone, APEX-EM A5 beats MemRL's CSR by +1.4pp (OS) and +1.0pp (DB).

\subsection{Embodied Navigation (ALFWorld)}

\begin{table}[t]
\centering
\resizebox{\linewidth}{!}{
\begin{tabular}{@{}llcc@{}}
\toprule
\textbf{Backbone} & \textbf{Method} & \textbf{Last SR} & \textbf{CSR} \\ \midrule
\multicolumn{4}{l}{\textit{MemRL baselines (GPT-5-mini)~\cite{memrl2025}}} \\
\midrule
GPT-5-mini & No Memory & 0.777 & -- \\
GPT-5-mini & Pass@10 & -- & 0.928 \\
GPT-5-mini & RAG & 0.907 & 0.935 \\
GPT-5-mini & Self-RAG & 0.907 & 0.962 \\
GPT-5-mini & Mem0 & 0.894 & 0.969 \\
GPT-5-mini & MemP & 0.885 & 0.919 \\
GPT-5-mini & MemRL & 0.949 & 0.981 \\
\midrule
\multicolumn{4}{l}{\textit{APEX-EM (ours, Claude Sonnet 4.5)}} \\
\midrule
Sonnet 4.5 & No Memory (A0) & 0.882 & 0.939 \\
Sonnet 4.5 & Semantic + judge (A2) & 0.998 & 0.998 \\
Sonnet 4.5 & Semantic + judge + iter.\ (A3) & 0.952 & 0.993 \\
Sonnet 4.5 & Structural (S1) & 0.965 & 0.991 \\
Sonnet 4.5 & Sem.\ + Struct.\ + Opus (A5) & \textbf{0.997} & \textbf{0.9996} \\
Sonnet 4.5 & Semantic + Structural (S2) & 0.955 & 0.992 \\
\midrule
\multicolumn{4}{l}{\textit{APEX-EM (ours, GPT-5-mini) — same backbone as MemRL}} \\
\midrule
GPT-5-mini & APEX-EM A5 (E1) & 0.991 & 0.999 \\
\bottomrule
\end{tabular}}
\caption{ALFWorld~\cite{shridhar2021alfworld} embodied navigation results. MemRL baselines from~\citet{memrl2025} with GPT-5-mini backbone. Last SR = final epoch success rate; CSR = cumulative success rate. A5 (Opus judge) achieves near-perfect 99.96\% CSR with Sonnet 4.5. GPT-5-mini APEX-EM achieves 99.1\% SR in first epoch, confirming saturation.}
\label{tab:alfworld_results}
\end{table}

APEX-EM essentially solves this benchmark: A5 achieves 99.7\% Last SR / 99.96\% CSR: only $\sim$1 task out of 2,511 is never solved. On the same GPT-5-mini backbone, APEX-EM A5 achieves 99.1\% SR in a single epoch, surpassing MemRL's 10-epoch result.

\section{HLE Ablation}
\label{app:hle_ablation}

\begin{table}[t]
\centering
\resizebox{\linewidth}{!}{
\begin{tabular}{@{}lcc@{}}
\toprule
\textbf{Configuration} & \textbf{HLE SR} & \textbf{CSR} \\ \midrule
A0: No Memory (1 epoch) & 25.2\% & -- \\
A1: Semantic, no judge & 19.4\% & 47.9\% \\
A3: Semantic + judge + iter.\ & 40.4\% & 52.9\% \\
R1: Semantic + rich + iter.\ & 45.6\% & 52.3\% \\
S2: Sem.\ + Structural & 46.8\% & 52.3\% \\
S1: Structural & \textbf{48.0\%} & \textbf{53.3\%} \\
\bottomrule
\end{tabular}}
\caption{HLE ablation study with Claude Opus 4.5 backbone. All ablations complete (10 epochs). S1 (structural retrieval) leads at 48.0\% SR / 53.3\% CSR; A1 (no judge, no iteration) shows high CSR (47.9\%) despite low SR (19.4\%), indicating broad exploration without convergence.}
\label{tab:hle_ablation}
\end{table}

Table~\ref{tab:hle_ablation} presents HLE-specific ablations with Opus 4.5, all complete at 10 epochs. The key finding is the stark contrast between A1 (semantic memory with binary reward, no judge, no iteration) at 19.4\% SR, versus the judged strategies with iteration (A3, R1, S1, S2) reaching 40.4--48.0\%. Structural retrieval (S1) leads at 48.0\% / 53.3\% CSR. Interestingly, A1 achieves high CSR (47.9\%) despite the lowest SR, indicating that binary reward enables broad exploration (solving different tasks each epoch) but not convergence (reliably solving the same tasks). All judged strategies converge to similar CSR (52--53\%), confirming that for expert-level reasoning, rich qualitative feedback is essential for consistent performance.

\section{HLE Backbone Scaling}
\label{app:hle_scaling}

\begin{table}[t]
\centering
\small
\resizebox{\linewidth}{!}{
\begin{tabular}{@{}lccc@{}}
\toprule
\textbf{Backbone} & \textbf{Baseline (A0)} & \textbf{APEX-EM S1 CSR} & \textbf{$\Delta$} \\ \midrule
Opus 4.5 & 25.2\% & 53.3\% & +28.1pp \\
Opus 4.6 & 36.6\% & 64.0\% & +27.4pp \\
Opus 4.7 & 39.4\% & 71.2\% & +31.8pp \\
\midrule
MemRL (Gemini-3-pro$^*$) & 35.7\% & 61.3\% & +25.6pp \\
\bottomrule
\end{tabular}}
\caption{HLE backbone scaling. APEX-EM S1 (structural retrieval) with three Opus checkpoints. CSR = tasks solved at least once across 10 epochs (frugal mode). $^*$Gemini-3-pro discontinued March 2026. Memory gain is additive to model capability (+27--32pp).}
\label{tab:hle_scaling}
\end{table}

To validate that APEX-EM's gains are not dependent on a specific backbone, we run the S1 (structural retrieval) configuration with three Claude Opus checkpoints of increasing capability. Table~\ref{tab:hle_scaling} shows that the memory gain is consistent (+27--32pp) across all three backbones, confirming that APEX-EM's benefit is additive to model capability rather than compensating for a specific model's weakness. We include MemRL's reported Gemini-3-pro result (61.3\% CSR) only as a historical reference: because that backbone was discontinued, we do \emph{not} treat this as a same-backbone comparison and do not claim the resulting gap as a controlled result. The controlled evidence here is the additive gain across the three Opus scales, not a head-to-head against MemRL.

\section{Iteration Efficiency}
\label{app:iteration_efficiency}

\begin{table}[t]
\centering
\resizebox{\linewidth}{!}{
\begin{tabular}{@{}lcccc@{}}
\toprule
\textbf{Config} & \textbf{E1 Avg} & \textbf{E10 Avg} & \textbf{$\Delta$} & \textbf{\% Red.} \\
\midrule
\multicolumn{5}{l}{\textit{BigCodeBench (max\_iterations=3)}} \\
\midrule
A3: Semantic + judge + iter. & 1.63 & 1.36 & $-$0.27 & 17\% \\
R1: Semantic + rich + iter. & 1.59 & 1.35 & $-$0.24 & 15\% \\
S1: Structural & 1.50 & 1.30 & $-$0.20 & 13\% \\
S2: Semantic + Structural & 1.50 & 1.32 & $-$0.18 & 12\% \\
A5: Sem.\ + Struct.\ + Opus & 1.51 & 1.35 & $-$0.16 & 11\% \\
\midrule
\multicolumn{5}{l}{\textit{KGQAGen-10k (max\_iterations=10)}} \\
\midrule
A5: Sem.\ + Struct.\ + Opus & 4.40 & 2.50 & $-$1.90 & 43\% \\
S2: Semantic + Structural & 4.42 & 2.73 & $-$1.69 & 38\% \\
R1: Semantic + rich + iter. & 4.19 & 2.61 & $-$1.58 & 38\% \\
A3: Semantic + judge + iter. & 4.12 & 2.59 & $-$1.53 & 37\% \\
S1: Structural & 4.02 & 2.52 & $-$1.50 & 37\% \\
\midrule
\multicolumn{5}{l}{\textit{Lifelong OS (max\_steps=15, agent turns)}} \\
\midrule
S1: Structural & 4.52 & 3.99 & $-$0.53 & 12\% \\
A3: Semantic + judge + iter. & 4.42 & 4.07 & $-$0.35 & 8\% \\
R1: Semantic + rich + iter. & 4.53 & 4.17 & $-$0.35 & 8\% \\
S2: Semantic + Structural & 4.44 & 4.10 & $-$0.33 & 8\% \\
A5: Sem.\ + Struct.\ + Opus & 4.42 & 4.17 & $-$0.25 & 6\% \\
\midrule
\multicolumn{5}{l}{\textit{ALFWorld (max\_steps=50, agent actions)}} \\
\midrule
A3: Semantic + judge + iter. & 9.53 & 7.91 & $-$1.62 & 17\% \\
A5: Sem.\ + Struct.\ + Opus & 9.04 & 7.80 & $-$1.24 & 14\% \\
\bottomrule
\end{tabular}}
\caption{Efficiency improvement from epoch 1 to epoch 10 (or last completed epoch). As memory accumulates, the agent requires fewer steps/iterations to solve tasks. BCB/KGQA measure self-correction iterations; LLB/ALFWorld measure agent environment steps. KGQA shows the largest reduction (37--43\%) due to high max\_iterations budget. ALFWorld shows 14--17\% step reduction despite near-saturated SR, indicating memory enables more direct solutions. LLB DB shows minimal reduction (2--4\%) as tasks are already solved in $\sim$3 steps.}
\label{tab:iteration_reduction}
\end{table}

Beyond improving success rates, memory accumulation reduces computational cost by decreasing self-correction iterations. On BCB (max\_iterations=3), iteration counts decrease by 11--17\% from E1 to E10. On KGQA (max\_iterations=10), the reduction is dramatic: 37--43\%, with A5 saving nearly 2 iterations per task on average (4.40$\to$2.50). On ALFWorld, step counts decrease by 14--17\% despite near-saturated success rates.

\begin{figure}[t]
  \centering
  \includegraphics[width=\linewidth]{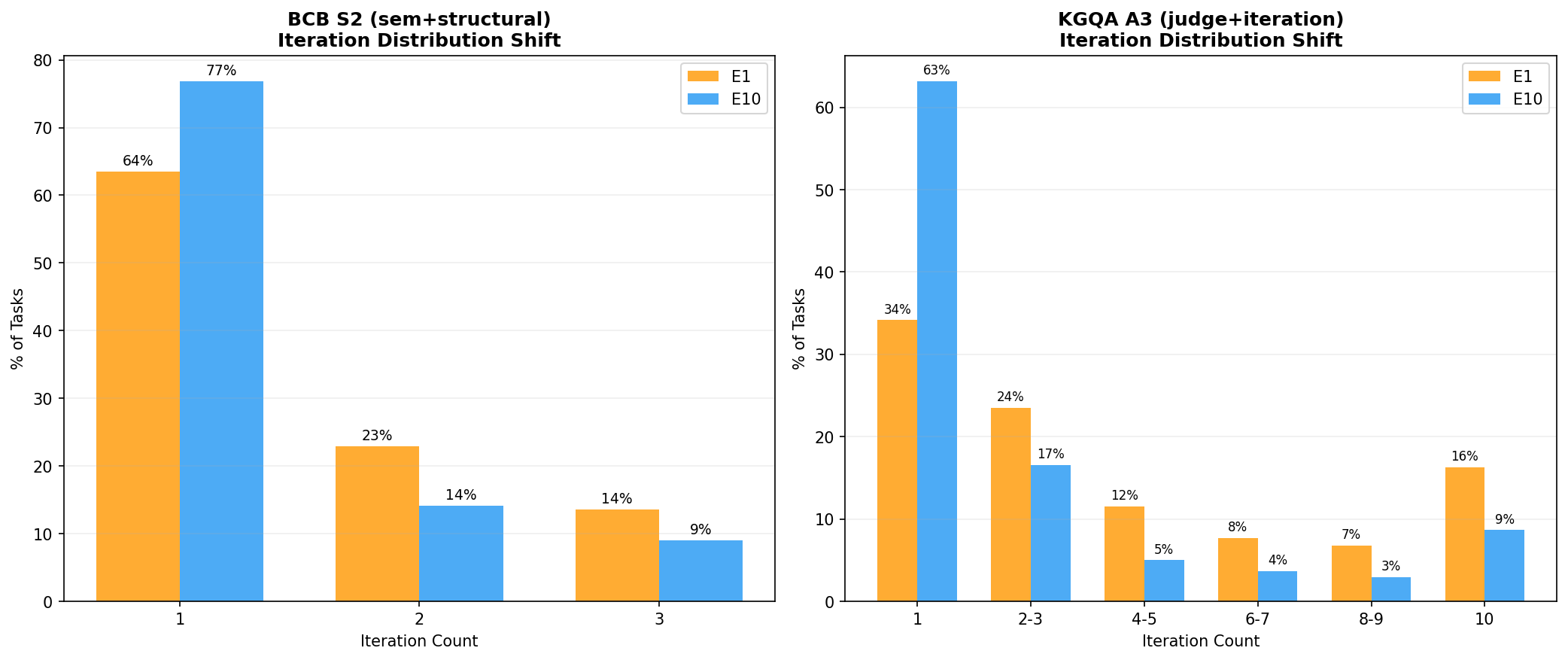}
  \caption{Iteration distribution shift from E1 to E10. \textbf{Left:} BCB S2, first-attempt solutions increase from 63.5\% to 76.8\% (+13.3pp). \textbf{Right:} KGQA A3, first-attempt solutions increase from 34.2\% to 63.1\% (+28.9pp).}
  \label{fig:iter_dist}
\end{figure}

The mechanism: in early epochs, iteration does most of the work (no memory yet); in later epochs, memory provides sufficient context that the initial generation is already correct, making iteration increasingly unnecessary.

\subsection{Cost per Correct Task}
\label{app:cost_analysis}

\begin{table}[t]
\centering
\small
\resizebox{\linewidth}{!}{
\begin{tabular}{@{}llcccccc@{}}
\toprule
& & \multicolumn{3}{c}{\textbf{Iterations}} & \multicolumn{3}{c}{\textbf{Cost per Correct Task}} \\
\cmidrule(lr){3-5} \cmidrule(lr){6-8}
\textbf{Benchmark} & \textbf{Method} & \textbf{E1} & \textbf{E10} & \textbf{$\Delta$\%} & \textbf{E1} & \textbf{E10} & \textbf{$\Delta$\%} \\
\midrule
\multirow{3}{*}{BCB}
& No Memory      & 1.63 & -- & -- & \$0.117 & -- & -- \\
& APEX-EM (S2)   & 1.50 & 1.32 & $-$12\% & \$0.108 & \$0.065 & $-$40\% \\
& APEX-EM (A5)   & 1.51 & 1.35 & $-$11\% & \$0.107 & \$0.064 & $-$40\% \\
\midrule
\multirow{3}{*}{KGQA}
& No Memory      & 4.12 & -- & -- & \$0.497 & -- & -- \\
& APEX-EM (S2)   & 4.42 & 2.73 & $-$38\% & \$0.304 & \$0.151 & $-$50\% \\
& APEX-EM (A5)   & 4.40 & 2.50 & $-$43\% & \$0.304 & \$0.137 & $-$55\% \\
\bottomrule
\end{tabular}}
\caption{Cost efficiency analysis. \textbf{Iterations}: average self-correction iterations per task. \textbf{Cost per Correct Task}: estimated LLM API cost amortized over successful tasks only, computed as (avg\_iters $\times$ cost\_per\_call) / success\_rate. Pricing: Sonnet~4.5 input \$3/MTok, output \$15/MTok; avg context ${\sim}$4K input + 1.5K output tokens per iteration. E1 = first epoch (no memory); E10 = tenth epoch (full memory). Cost reduction combines both iteration reduction and success rate improvement.}
\label{tab:cost_analysis}
\end{table}

Table~\ref{tab:cost_analysis} quantifies the cost efficiency gains from memory accumulation by computing the amortized LLM API cost per \emph{correctly solved} task. We estimate per-iteration cost using Sonnet~4.5 pricing (\$3/MTok input, \$15/MTok output) with average context sizes of ${\sim}$4K input + 1.5K output tokens for BCB and ${\sim}$5K input + 1K output tokens for KGQA, yielding approximately \$0.035/iteration (BCB) and \$0.030/iteration (KGQA). The cost per correct task is then: $\text{cost}_{\text{correct}} = (\bar{J} \times c_{\text{iter}}) / \text{SR}$, where $\bar{J}$ is the average iteration count and SR is the success rate.

The dual benefit of memory is evident: iteration reduction (\emph{fewer calls per task}) combines multiplicatively with success rate improvement (\emph{fewer wasted attempts}), yielding 40--55\% cost reduction by epoch 10. The retrieval overhead itself is negligible: Titan Embed V2 at \$0.02/MTok adds $<$\$0.001/query ($<$1\% of iteration cost), and graph traversal is a local operation with sub-millisecond latency. This demonstrates that APEX-EM's memory overhead is amortized within the first few tasks of each epoch.

\section{Implementation Architecture}
\label{app:architecture}

The \texttt{PRGIIWorkflow} class orchestrates the five phases via dependency injection. It accepts a \texttt{PRGIIConfig} (phase toggles and parameters) and five phase objects. The \texttt{execute()} method runs phases in sequence: Plan $\to$ Retrieve $\to$ [Generate $\leftrightarrow$ Iterate]$^J$ $\to$ Ingest. Each phase is skipped when disabled, enabling ablation studies by toggling individual components.

Four extension points enable domain specialization: (1) \textbf{DomainAdapter Protocol}: defines entity/property/procedural-step extraction methods; (2) \textbf{PromptRegistry}: maps task types to domain-specific prompts for each phase; (3) \textbf{Validators}: domain-specific verification for Iteration ($\mathcal{V}_d$) and Ingest ($\mathcal{O}_d$) phases; (4) \textbf{PRGIIConfig}: boolean toggles for each phase plus max\_iterations and quality\_threshold.

\begin{table}[h]
\centering
\resizebox{\linewidth}{!}{
\begin{tabular}{@{}lccccc@{}}
\toprule
\textbf{ID} & \textbf{Retrieval} & \textbf{Iterate} & \textbf{Judge} & \textbf{Feedback} & \textbf{Judge Model} \\ \midrule
A0 & --- & --- & --- & --- & --- \\
A1 & Semantic & --- & --- & Binary & Sonnet \\
A2 & Semantic & --- & $\checkmark$ & Rich & Sonnet \\
A3 & Semantic & $\checkmark$ & $\checkmark$ & Binary & Sonnet \\
R1 & Semantic & $\checkmark$ & $\checkmark$ & Rich & Sonnet \\
S1 & Structural & $\checkmark$ & $\checkmark$ & Rich & Sonnet \\
S2 & Semantic + Structural & $\checkmark$ & $\checkmark$ & Rich & Sonnet \\
A5 & Semantic + Structural & $\checkmark$ & $\checkmark$ & Rich & Opus \\
\bottomrule
\end{tabular}}
\caption{Ablation configurations. \textbf{Retrieval}: which retrieval channels are active (Semantic = embedding similarity; Structural = signature matching via entity-centric index; Semantic + Structural = both). \textbf{Iterate}: iterative refinement on failed tasks. \textbf{Judge}: quality judge on ingested plans. \textbf{Feedback}: what the agent sees from retrieved plans (Binary = pass/fail only; Rich = quality scores + rationale). \textbf{Judge Model}: LLM used for quality assessment (Sonnet = Claude Sonnet 4.5; Opus = Claude Opus 4.5). All configs with retrieval also include planning and ingest phases.}
\label{tab:strategy_configs}
\end{table}

\section{Entity Graph Visualizations}
\label{app:entity_graph_viz}

Figures~\ref{fig:eg_bcb}--\ref{fig:eg_hle} illustrate Entity Graphs from each benchmark domain. All share the same 4 node types and 5 edge types but with domain-specific instantiations.

\begin{figure*}[t]
\centering
\resizebox{\textwidth}{!}{%
\begin{tikzpicture}[
  entity/.style={rectangle, rounded corners=4pt, draw=blue!70, fill=blue!8, font=\sffamily\small, minimum height=0.7cm, inner sep=4pt},
  property/.style={rectangle, rounded corners=4pt, draw=orange!70, fill=orange!8, font=\sffamily\small, minimum height=0.7cm, inner sep=4pt},
  procedural/.style={rectangle, rounded corners=4pt, draw=red!60, fill=red!8, font=\sffamily\small, minimum height=0.7cm, inner sep=4pt},
  topic/.style={ellipse, draw=green!60!black, fill=green!10, font=\sffamily\small\bfseries, minimum height=0.8cm, inner sep=4pt},
  arrow/.style={-{Stealth[length=4pt]}, semithick},
  edgelabel/.style={font=\sffamily\tiny, fill=white, inner sep=1pt},
]
\node[topic] (topic) at (7, 4.5) {Data Analysis};
\node[entity] (pandas) at (2, 2.5) {pandas};
\node[entity] (df) at (5, 2.5) {DataFrame};
\node[entity] (csv) at (0, 0.5) {csv};
\node[entity] (numpy) at (9.5, 2.5) {numpy};
\node[entity] (matplotlib) at (12, 2.5) {matplotlib};
\node[property] (groupby) at (2, 0.5) {groupby};
\node[property] (read_csv) at (4.5, 0.5) {read\_csv};
\node[property] (rolling) at (7, 0.5) {rolling};
\node[property] (mean) at (9.5, 0.5) {mean};
\node[property] (plot) at (12, 0.5) {plot};
\node[procedural] (load) at (1.5, -1.8) {data\_loading};
\node[procedural] (transform) at (5, -1.8) {transformation};
\node[procedural] (agg) at (8.5, -1.8) {aggregation};
\node[procedural] (viz) at (12, -1.8) {visualization};
\draw[arrow, green!50!black, densely dotted] (pandas) -- (topic) node[edgelabel, pos=0.5] {\textsc{member\_of}};
\draw[arrow, green!50!black, densely dotted] (df) -- (topic);
\draw[arrow, green!50!black, densely dotted] (numpy) -- (topic);
\draw[arrow, green!50!black, densely dotted] (matplotlib) -- (topic);
\draw[arrow, orange!70] (pandas) -- (groupby) node[edgelabel, pos=0.5, left] {\textsc{relates\_to}};
\draw[arrow, orange!70] (pandas) -- (read_csv);
\draw[arrow, orange!70] (df) -- (rolling);
\draw[arrow, orange!70] (df) -- (groupby);
\draw[arrow, orange!70] (numpy) -- (mean);
\draw[arrow, orange!70] (matplotlib) -- (plot);
\draw[arrow, red!60, thick] (load) to[out=-20, in=-160] node[edgelabel, pos=0.5, below=2pt] {\textsc{followed\_by}} (transform);
\draw[arrow, red!60, thick] (transform) to[out=-20, in=-160] node[edgelabel, pos=0.5, below=2pt] {\textsc{followed\_by}} (agg);
\draw[arrow, red!60, thick] (agg) to[out=-20, in=-160] node[edgelabel, pos=0.5, below=2pt] {\textsc{followed\_by}} (viz);
\draw[arrow, purple!60, dashed] (load) -- (read_csv) node[edgelabel, pos=0.4] {\textsc{uses}};
\draw[arrow, purple!60, dashed] (transform) -- (groupby);
\draw[arrow, purple!60, dashed] (agg) -- (rolling);
\draw[arrow, purple!60, dashed] (agg) -- (mean);
\draw[arrow, purple!60, dashed] (viz) -- (plot);
\node[draw=black!30, rounded corners=3pt, fill=yellow!5, font=\sffamily\scriptsize, align=left, inner sep=4pt]
  (taskA) at (1.5, -3.5) {\textbf{Task A} $\hat{\sigma}$: data\_loading $\to$ transformation $\to$ aggregation};
\node[draw=black!30, rounded corners=3pt, fill=yellow!5, font=\sffamily\scriptsize, align=left, inner sep=4pt]
  (taskB) at (9, -3.5) {\textbf{Task B} $\hat{\sigma}$: data\_loading $\to$ transformation $\to$ aggregation $\to$ visualization};
\draw[densely dotted, black!40] (taskA.north) -- (load.south);
\draw[densely dotted, black!40] (taskB.north) -- (viz.south);
\end{tikzpicture}%
}
\caption{BCB Entity Graph. Tasks A and B share entities and procedural steps, enabling structural retrieval despite different descriptions.}
\label{fig:eg_bcb}
\end{figure*}

\begin{figure*}[t]
\centering
\resizebox{\textwidth}{!}{%
\begin{tikzpicture}[
  entity/.style={rectangle, rounded corners=4pt, draw=blue!70, fill=blue!8, font=\sffamily\small, minimum height=0.7cm, inner sep=4pt},
  property/.style={rectangle, rounded corners=4pt, draw=orange!70, fill=orange!8, font=\sffamily\small, minimum height=0.7cm, inner sep=4pt},
  procedural/.style={rectangle, rounded corners=4pt, draw=red!60, fill=red!8, font=\sffamily\small, minimum height=0.7cm, inner sep=4pt},
  topic/.style={ellipse, draw=green!60!black, fill=green!10, font=\sffamily\small\bfseries, minimum height=0.8cm, inner sep=4pt},
  arrow/.style={-{Stealth[length=4pt]}, semithick},
  edgelabel/.style={font=\sffamily\tiny, fill=white, inner sep=1pt},
]
\node[topic] (topic) at (7, 4.5) {Comparative Statistics};
\node[entity] (curry) at (0.5, 2.5) {Stephen Curry (Q352159)};
\node[entity] (nba) at (4.5, 2.5) {NBA (Q155223)};
\node[entity] (season) at (8.5, 2.5) {2023--24 season (Q...)};
\node[entity] (lebron) at (12.5, 2.5) {LeBron James (Q36159)};
\node[property] (pts) at (1, 0.5) {points scored (P1351)};
\node[property] (team) at (4.5, 0.5) {member of (P54)};
\node[property] (season_p) at (8, 0.5) {season (P3450)};
\node[property] (assists) at (11.5, 0.5) {assists (P5765)};
\node[procedural] (eres) at (0.5, -1.8) {entity\_resolution};
\node[procedural] (schema) at (4, -1.8) {schema\_traversal};
\node[procedural] (temporal) at (7.5, -1.8) {temporal\_filter};
\node[procedural] (agg) at (10.5, -1.8) {aggregation};
\node[procedural] (comp) at (13.5, -1.8) {comparison};
\draw[arrow, green!50!black, densely dotted] (curry) -- (topic);
\draw[arrow, green!50!black, densely dotted] (nba) -- (topic);
\draw[arrow, green!50!black, densely dotted] (lebron) -- (topic);
\draw[arrow, orange!70] (curry) -- (pts) node[edgelabel, pos=0.5, left] {\textsc{relates\_to}};
\draw[arrow, orange!70] (curry) -- (team);
\draw[arrow, orange!70] (nba) -- (season_p);
\draw[arrow, orange!70] (lebron) -- (assists);
\draw[arrow, orange!70] (lebron) -- (pts);
\draw[arrow, cyan!60, thick, densely dash dot] (curry) to[bend left=20] node[edgelabel, pos=0.5, above] {\textsc{equivalent\_to}} (lebron);
\draw[arrow, red!60, thick] (eres) to[out=-20, in=-160] node[edgelabel, pos=0.5, below=2pt] {\textsc{followed\_by}} (schema);
\draw[arrow, red!60, thick] (schema) to[out=-20, in=-160] node[edgelabel, pos=0.5, below=2pt] {\textsc{followed\_by}} (temporal);
\draw[arrow, red!60, thick] (temporal) to[out=-20, in=-160] node[edgelabel, pos=0.5, below=2pt] {\textsc{followed\_by}} (agg);
\draw[arrow, red!60, thick] (agg) to[out=-20, in=-160] node[edgelabel, pos=0.5, below=2pt] {\textsc{followed\_by}} (comp);
\draw[arrow, purple!60, dashed] (eres) -- (team);
\draw[arrow, purple!60, dashed] (schema) -- (pts) node[edgelabel, pos=0.3] {\textsc{uses}};
\draw[arrow, purple!60, dashed] (temporal) -- (season_p);
\draw[arrow, purple!60, dashed] (agg) -- (assists);
\node[draw=black!30, rounded corners=3pt, fill=yellow!5, font=\sffamily\scriptsize, align=left, inner sep=4pt]
  (taskC) at (1, -3.5) {\textbf{Task C}: ``Curry's 3-pointers vs last season''\\$\hat{\sigma}$: eres $\to$ schema $\to$ temporal $\to$ agg $\to$ comp};
\node[draw=black!30, rounded corners=3pt, fill=yellow!5, font=\sffamily\scriptsize, align=left, inner sep=4pt]
  (taskD) at (10, -3.5) {\textbf{Task D}: ``Compare LeBron's assists across seasons''\\$\hat{\sigma}$: eres $\to$ schema $\to$ temporal $\to$ agg $\to$ comp};
\draw[densely dotted, black!40] (taskC.north) -- (eres.south);
\draw[densely dotted, black!40] (taskD.north) -- (comp.south);
\end{tikzpicture}%
}
\caption{KGQA Entity Graph. Tasks C and D produce identical structural signatures despite different entities, enabling cross-entity structural retrieval.}
\label{fig:eg_kgqa}
\end{figure*}

\begin{figure*}[t]
\centering
\resizebox{\textwidth}{!}{%
\begin{tikzpicture}[
  entity/.style={rectangle, rounded corners=4pt, draw=blue!70, fill=blue!8, font=\sffamily\small, minimum height=0.7cm, inner sep=4pt},
  property/.style={rectangle, rounded corners=4pt, draw=orange!70, fill=orange!8, font=\sffamily\small, minimum height=0.7cm, inner sep=4pt},
  procedural/.style={rectangle, rounded corners=4pt, draw=red!60, fill=red!8, font=\sffamily\small, minimum height=0.7cm, inner sep=4pt},
  topic/.style={ellipse, draw=green!60!black, fill=green!10, font=\sffamily\small\bfseries, minimum height=0.8cm, inner sep=4pt},
  arrow/.style={-{Stealth[length=4pt]}, semithick},
  edgelabel/.style={font=\sffamily\tiny, fill=white, inner sep=1pt},
]
\node[topic] (topic) at (7, 4.5) {Algebraic Topology};
\node[entity] (spin) at (0.5, 2.5) {Spin bordism};
\node[entity] (g2) at (4, 2.5) {$G_2$ (Lie group)};
\node[entity] (bg2) at (7.5, 2.5) {$BG_2$ (classifying space)};
\node[entity] (ahss) at (11.5, 2.5) {Atiyah-Hirzebruch SS};
\node[property] (homology) at (1, 0.5) {integral homology};
\node[property] (cohoring) at (4.5, 0.5) {cohomology ring};
\node[property] (spectral) at (8, 0.5) {spectral sequence};
\node[property] (ext) at (11.5, 0.5) {Ext groups};
\node[procedural] (formulate) at (0.5, -1.8) {problem\_formulation};
\node[procedural] (derive) at (4, -1.8) {mathematical\_derivation};
\node[procedural] (compute) at (8, -1.8) {numerical\_computation};
\node[procedural] (verify) at (12, -1.8) {verification};
\draw[arrow, green!50!black, densely dotted] (spin) -- (topic);
\draw[arrow, green!50!black, densely dotted] (g2) -- (topic);
\draw[arrow, green!50!black, densely dotted] (bg2) -- (topic);
\draw[arrow, green!50!black, densely dotted] (ahss) -- (topic);
\draw[arrow, orange!70] (spin) -- (homology) node[edgelabel, pos=0.5, left] {\textsc{relates\_to}};
\draw[arrow, orange!70] (bg2) -- (cohoring);
\draw[arrow, orange!70] (ahss) -- (spectral);
\draw[arrow, orange!70] (ahss) -- (ext);
\draw[arrow, orange!70] (g2) -- (cohoring);
\draw[arrow, red!60, thick] (formulate) to[out=-20, in=-160] node[edgelabel, pos=0.5, below=2pt] {\textsc{followed\_by}} (derive);
\draw[arrow, red!60, thick] (derive) to[out=-20, in=-160] node[edgelabel, pos=0.5, below=2pt] {\textsc{followed\_by}} (compute);
\draw[arrow, red!60, thick] (compute) to[out=-20, in=-160] node[edgelabel, pos=0.5, below=2pt] {\textsc{followed\_by}} (verify);
\draw[arrow, purple!60, dashed] (formulate) -- (homology) node[edgelabel, pos=0.3] {\textsc{uses}};
\draw[arrow, purple!60, dashed] (derive) -- (spectral);
\draw[arrow, purple!60, dashed] (derive) -- (cohoring);
\draw[arrow, purple!60, dashed] (compute) -- (ext);
\node[draw=black!30, rounded corners=3pt, fill=yellow!5, font=\sffamily\scriptsize, align=left, inner sep=4pt]
  (taskE) at (1, -3.5) {\textbf{Task E}: ``Compute $\tilde{\Omega}_{12}^{Spin}(BG_2)$''\\$\hat{\sigma}$: formulation $\to$ derivation $\to$ computation $\to$ verification};
\node[draw=black!30, rounded corners=3pt, fill=yellow!5, font=\sffamily\scriptsize, align=left, inner sep=4pt]
  (taskF) at (9.5, -3.5) {\textbf{Task F}: ``Compute $\pi_4(S^3)$ via AHSS''\\$\hat{\sigma}$: formulation $\to$ derivation $\to$ computation $\to$ verification};
\draw[densely dotted, black!40] (taskE.north) -- (formulate.south);
\draw[densely dotted, black!40] (taskF.north) -- (verify.south);
\end{tikzpicture}%
}
\caption{HLE Entity Graph. Tasks E and F share the same procedural pattern and the Atiyah-Hirzebruch spectral sequence entity, enabling structural retrieval across different algebraic topology problems.}
\label{fig:eg_hle}
\end{figure*}

\section{Judge Feedback Without Answer Leakage}
\label{app:judge_feedback}

A critical design property of APEX-EM is that the Judge LLM evaluates the agent's \emph{proof-of-work} (the execution trace, reasoning chain, and generated artifacts) without leaking the ground-truth answer into the feedback stored in memory. The judge has access to the oracle answer for scoring, but the feedback text identifies \emph{where the reasoning went wrong} rather than \emph{what the correct answer is}.

\paragraph{Example 1: Spin Bordism (Algebraic Topology).}
\emph{Task}: Compute the reduced 12th dimensional Spin bordism of the classifying space of $G_2$. \emph{Gold}: $\mathbb{Z}^5$.

In epoch 1, the agent computed $\mathbb{Z}^3$ using the correct Atiyah-Hirzebruch spectral sequence approach but with incorrect homology groups. The judge feedback stated: ``\emph{homology groups of $BG_2$ are computed incorrectly}'' and directed: ``\emph{carefully compute the integral homology using the known cohomology ring structure.}'' The feedback identifies the specific computational step that failed without revealing $\mathbb{Z}^5$. In epoch 2, with this failed experience retrieved as a negative example, the agent arrived at the correct answer.

\paragraph{Example 2: Multi-Part Cipher Puzzle.}
\emph{Task}: A multi-part puzzle involving logical depth, a Gell-Mann quote, GELU authorship, ROT13 cipher, and planetary mass comparison. \emph{Gold}: ``yeyo''.

The judge feedback identified: ``\emph{identification of `crypticity' as the reciprocal concept needs verification}'' and ``\emph{the quote about MIT and options needs more careful analysis.}'' The feedback pinpointed which sub-parts were wrong without revealing the final answer.

\paragraph{Example 3: Elliptic Curve Torsion (Number Theory).}
\emph{Task}: Largest order of a non-cyclic torsion subgroup of an elliptic curve over $\mathbb{Q}(\sqrt{-3})$. \emph{Gold}: 18.

The agent claimed $\mathbb{Z}/2\mathbb{Z} \times \mathbb{Z}/14\mathbb{Z}$ (order 28). The judge feedback stated: ``\emph{incorrectly claims $\mathbb{Z}/2\mathbb{Z} \times \mathbb{Z}/14\mathbb{Z}$ occurs without proper verification}'' and directed: ``\emph{consult the specific classification results for $\mathbb{Q}(\sqrt{-3})$ more carefully.}'' In epoch 2, the agent correctly identified $\mathbb{Z}/3\mathbb{Z} \times \mathbb{Z}/6\mathbb{Z}$ (order 18).

\section{Architectural Comparison with MemRL}
\label{app:comparison_note}

Two architectural differences beyond the memory representation distinguish APEX-EM from MemRL: (1) APEX-EM makes validation an explicit phase within the iteration loop, where the agent self-verifies each attempt against domain-specific criteria ($\mathcal{V}_d$), producing structured feedback that guides the next attempt, rather than relying solely on implicit LLM self-assessment; and (2) the Teacher model (Opus 4.5) independently evaluates completed traces at the commit phase (Ingest), providing multi-dimensional feedback that determines whether experiences enter memory. The ablation in Table~\ref{tab:ablation} isolates these contributions: A1 (memory with binary reward, no teacher) shows gains attributable purely to memory retrieval, the A1$\to$A3 gap quantifies explicit self-verification, and the A2$\to$A5 gap quantifies teacher-quality feedback at commit.

\section{Self-Improvement Mechanisms}
\label{app:self_improvement}

PRGII improves over single-shot generation through two independent mechanisms:

\paragraph{Mechanism 1: Cross-Epoch Memory Retrieval.}
BCB task 15 (execute shell commands from CSV, save outputs to files): the single-shot LLM generates reasonable code but names output files incorrectly. The oracle test suite expects a different naming convention that cannot be inferred from the task description alone. In epoch 2, the failed experience from epoch 1 is retrieved as a negative example with the oracle's error message, and the agent reverse-engineers the correct naming convention.

\paragraph{Mechanism 2: Within-Epoch Self-Correction Iteration.}
BCB task 89 (outlier removal via Z-score with StandardScaler): the LLM generates code using strict inequality (\texttt{>}) where the oracle expects inclusive (\texttt{>=}). Over 3 iterations: (1) code generated with \texttt{>}, (2) reviewer catches the discrepancy, (3) code updated to \texttt{>=}, tests pass. The self-critique loop catches a subtle off-by-one bug without any memory involvement.

These mechanisms are complementary: memory retrieval transfers knowledge across epochs, while iteration catches bugs within a single execution. The ablation results confirm this: A1 (memory, no iteration) and A3 (memory + iteration) both improve over A0, with gains partially additive in the full pipeline.

\paragraph{Why Iteration Enables KGQA Transfer.}
Consider a training question about Stephen Curry's statistics and a transfer question about LeBron James's career requiring the same properties but different entity QIDs. Memory retrieval surfaces the Curry example as structurally similar. With iteration, the agent can generate an initial query using the retrieved pattern, execute it, and refine when results are unexpected. Without iteration, the agent must get the query right on the first attempt, and subtle differences between entities often cause failure. This explains the stark A1/A2 vs A3+ gap in transfer results.

\section{Reproducibility and Artifacts}
\label{app:code_release}

We will release the following artifacts to reproduce the study:
\begin{itemize}
\item \textbf{Code.} The core APEX-EM implementation: the PRGII workflow orchestrator, PKG construction and retrieval modules (semantic, structural-signature, and graph-traversal channels), structural-signature extraction, and the domain-adapter interfaces.
\item \textbf{PKG schema.} The full Experience ontology and five node / ten edge types (Table~\ref{tab:experience-ontology}, \S\ref{sec:pkg}), released as a machine-readable schema.
\item \textbf{Prompts.} The domain-specific Plan/Generate/Iterate/Ingest prompt templates for all five benchmarks (\S\ref{app:pipelines}), plus the Teacher-judge and EntityResolver prompts.
\item \textbf{Models and decoding.} Backbones (Claude Sonnet 4.5, Opus 4.5/4.6/4.7; GPT-4o, GPT-4o-mini, GPT-5-mini; Amazon Titan Embed v2) with exact version identifiers, temperature 0, and the quality-gate/threshold settings ($\theta=0.3$, $w_c=0.9$), all listed in the released configuration.
\item \textbf{Splits and harnesses.} The MemRL-compatible BCB 798/342 train/transfer split (seed 42), the KGQAGen-10k 2{,}000-train / 1{,}079-test split, and the HLE ablation subset (1{,}000-q, seed 42), with per-benchmark evaluation harnesses.
\end{itemize}
Because every state-of-the-art baseline we compare against reports on closed-weight models, the main results use closed weights for a faithful matched comparison; this release lets the pipeline be reproduced independently of any single API model. All benchmarks are used under their respective licenses; no new data was collected and no personally identifying information is involved.

\section{Retrieval-Channel Isolation}
\label{app:channel_isolation}

Reviewers asked us to decouple the three retrieval channels rather than bundle them. Table~\ref{tab:channel_isolation} isolates semantic search, structural-signature matching, and graph traversal, along with their combinations, under one protocol (last-epoch SR, training split; HLE on the 1{,}000-question ablation subset, Opus 4.5).

\begin{table}[t]
\centering
\small
\begin{tabular}{@{}lccc@{}}
\toprule
\textbf{Retrieval channel} & \textbf{BCB} & \textbf{KGQA} & \textbf{HLE} \\
\midrule
Semantic only (R1)          & 80.1 & 85.7 & 45.6 \\
Structural only (S1)        & 78.6 & 88.5 & 48.0 \\
Graph traversal only (EG1)  & 78.6 & 85.5 & 48.0 \\
Semantic + Structural (S2)  & 81.1 & 90.6 & 48.8 \\
All three (A5)              & \textbf{83.3} & \textbf{91.2} & \textbf{48.8} \\
\bottomrule
\end{tabular}
\caption{Retrieval-channel isolation (last-epoch SR, training split; HLE on the 1{,}000-q ablation subset, Opus 4.5). On KGQA the structural signature is the strongest single channel (+2.8\,pp over semantic); graph traversal adds recall on multi-hop links; the three are additive. On code the channels are close because execution feedback already resolves most tasks.}
\label{tab:channel_isolation}
\end{table}

\section{Scoring-Weight Sensitivity}
\label{app:sensitivity}

We sweep the composite ranking weights $(\alpha, \beta, \gamma, \delta, \epsilon)$ of Eq.~\ref{eq:ranking} (semantic, structural, graph-proximity, quality, recency). The sweep uses 50 sampled tasks per benchmark, so absolute numbers differ slightly from the full-set tables. Row~1 is the paper default.

\begin{table}[t]
\centering
\small
\resizebox{\linewidth}{!}{
\begin{tabular}{@{}lccccccc@{}}
\toprule
\textbf{Config} & $\alpha$ & $\beta$ & $\gamma$ & $\delta$ & $\epsilon$ & \textbf{BCB} & \textbf{KGQA} \\
\midrule
Default          & 0.40 & 0.30 & 0.20 & 0.10 & 0.10 & \textbf{82.4} & \textbf{90.6} \\
Semantic-heavy   & 0.60 & 0.20 & 0.10 & 0.05 & 0.05 & 80.2 & 86.2 \\
Structure-heavy  & 0.20 & 0.50 & 0.20 & 0.05 & 0.05 & 79.6 & 90.6 \\
Graph-heavy      & 0.20 & 0.20 & 0.50 & 0.05 & 0.05 & 78.6 & 89.5 \\
Uniform          & 0.20 & 0.20 & 0.20 & 0.20 & 0.20 & 76.4 & 87.4 \\
No recency       & 0.45 & 0.30 & 0.20 & 0.05 & 0.00 & 77.6 & 86.4 \\
No quality prior & 0.45 & 0.30 & 0.25 & 0.00 & 0.00 & 74.7 & 82.4 \\
\bottomrule
\end{tabular}}
\caption{Ranking-weight sensitivity (aggregate SR over 50 sampled tasks per benchmark; HLE SR follows the same pattern, 49.2 default down to 44.8 without the quality prior). Performance is robust across reasonable weightings and degrades only at extremes. The quality prior is the single most important term (dropping it costs the most: 74.7/82.4), keeping proven procedures ranked above merely relevant ones; recency is next, retaining the latest successful procedure once a task is solved. The balance among recall channels is task-dependent: semantic-heavy weighting hurts structured tasks most (KGQA 90.6$\rightarrow$86.2), while code stays closer because execution feedback resolves most tasks. This motivates a learned per-task merge policy as future work.}
\label{tab:sensitivity}
\end{table}

\paragraph{Minimal configuration per domain.}
The smallest component set that recovers at least 95\% of full-pipeline performance differs by domain: for \textbf{code} (BCB), neither the teacher nor structural signatures are needed (semantic retrieval with iteration suffices, since execution feedback is decisive); for \textbf{structured queries} (KGQA/SPARQL), the teacher is required (rich feedback contributes +10.3\,pp); and for \textbf{cross-domain transfer}, structural signatures plus iteration are the minimal set. This per-domain split is the basis for the modular-memory recommendation.

\section{Latency Analysis}
\label{app:latency}

Latency tracks iteration count, because each iteration is one generate-then-validate round trip; retrieval adds little (graph traversal is sub-millisecond and embedding lookup is under \$0.001 per query, below 1\% of one iteration). Table~\ref{tab:latency} pairs the iteration reduction with an end-to-end latency view of the same runs.

\begin{table}[t]
\centering
\small
\resizebox{\linewidth}{!}{
\begin{tabular}{@{}llcccc@{}}
\toprule
\textbf{Bench} & \textbf{Cfg} & \textbf{Iters E1$\to$E10} & \textbf{Lat.\ $\Delta$} & \textbf{1st-try E1$\to$E10} & \textbf{Latency/task} \\
\midrule
BCB  & S2 & 1.50 $\to$ 1.32 & $-$12\% & 63.5\% $\to$ 76.8\% & 420\,s $\to$ 370\,s \\
KGQA & A3 & 4.40 $\to$ 2.50 & $-$43\% & 34.2\% $\to$ 63.1\% & 380\,s $\to$ 215\,s \\
\bottomrule
\end{tabular}}
\caption{Latency view (iteration-driven; retrieval/graph overhead $<$1\,ms traversal and $<$\$0.001 embedding per query). End-to-end wall-clock per task drops as first-attempt solve rate rises with memory; these figures are estimated from run logs.}
\label{tab:latency}
\end{table}

\section{Failure Modes}
\label{app:failure_modes}

\begin{table}[t]
\centering
\small
\resizebox{\linewidth}{!}{
\begin{tabular}{@{}llll@{}}
\toprule
\textbf{Failure mode} & \textbf{Domain} & \textbf{Freq.} & \textbf{Handling} \\
\midrule
Degenerate single-op sig. & all  & $<$2\% & fall back to semantic-only retrieval \\
Entity-resolution miss     & KGQA & $<$4\% & new node created; can weaken match \\
Sandbox/exec.\ timeout     & BCB  & $<$3\% & scored as failure; error registry filled \\
Near-miss template reused  & all  & $<$6\% & failed-plan ($P^-$) guardrail catches it \\
\bottomrule
\end{tabular}}
\caption{Per-domain failure taxonomy. Structural-signature extraction failures (degenerate single-operation signatures) occur in under 2\% of tasks and fall back gracefully to semantic-only retrieval.}
\label{tab:failure_modes}
\end{table}

\end{document}